\documentclass[10pt,twocolumn,dvipsnames]{article}

\makeatletter
\renewcommand*{\@fnsymbol}[1]{\ifnum#1=1*\else\@arabic{\numexpr#1-1\relax}\fi}
\makeatother

\usepackage{cvpr}              % To produce the CAMERA-READY version
\usepackage{mathrsfs}
\usepackage{pifont} % Add this in your preamble

\usepackage{graphicx}
\usepackage[dvipsnames]{xcolor}

\usepackage{amsmath,amsfonts,bm}

\def\eqref#1{equation~\ref{#1}}
\def\1{\bm{1}}

\def\rmB{{\mathbf{B}}}
\def\rmC{{\mathbf{C}}}

\def\rmH{{\mathbf{H}}}
\def\rmI{{\mathbf{I}}}

\def\rmK{{\mathbf{K}}}
\def\rmL{{\mathbf{L}}}
\def\rmM{{\mathbf{M}}}

\def\rmP{{\mathbf{P}}}
\def\rmQ{{\mathbf{Q}}}
\def\rmR{{\mathbf{R}}}
\def\rmS{{\mathbf{S}}}

\def\rmW{{\mathbf{W}}}
\def\rmX{{\mathbf{X}}}
\def\rmY{{\mathbf{Y}}}
\def\rmZ{{\mathbf{Z}}}

\def\vone{{\bm{1}}}

\def\vb{{\bm{b}}}
\def\vc{{\bm{c}}}

\def\vf{{\bm{f}}}
\def\vg{{\bm{g}}}
\def\vh{{\bm{h}}}

\def\vq{{\bm{q}}}

\def\vv{{\bm{v}}}

\def\vx{{\bm{x}}}

\def\vz{{\bm{z}}}

\DeclareMathAlphabet{\mathsfit}{\encodingdefault}{\sfdefault}{m}{sl}
\SetMathAlphabet{\mathsfit}{bold}{\encodingdefault}{\sfdefault}{bx}{n}

\def\gB{{\mathcal{B}}}
\def\gC{{\mathcal{C}}}

\def\gO{{\mathcal{O}}}
\def\gP{{\mathcal{P}}}
\def\gQ{{\mathcal{Q}}}

\def\gX{{\mathcal{X}}}

\def\sR{{\mathbb{R}}}

\def\emR{{R}}

\DeclareMathOperator*{\argmax}{arg\,max}
\DeclareMathOperator*{\argmin}{arg\,min}

\DeclareMathOperator{\tr}{tr}

\DeclareMathOperator{\cka}{CKA}
\DeclareMathOperator{\localcka}{localCKA}
\DeclareMathOperator{\hsic}{HSIC}

\usepackage[dvipsnames]{xcolor}
\usepackage{color, colortbl}
\usepackage{nicefrac}

\usepackage{array}
\usepackage{adjustbox}
\usepackage{array, makecell}

\definecolor{Gray}{gray}{0.9}
\definecolor{brightturquoise}{rgb}{0.85, 1, 1}
\definecolor{newpurple}{HTML}{BC61F5}%EF9B0F
\newcommand{\mycc}{\cellcolor[HTML]{EDF6FF}}

\definecolor{cvprblue}{rgb}{0.21,0.49,0.74}
\usepackage[pagebackref,breaklinks,colorlinks,citecolor=cvprblue]{hyperref}
\usepackage{booktabs}
\usepackage{rotating}

\usepackage{multirow}
\newcommand{\cmark}{\ding{51}}%
\newcommand{\xmark}{\ding{55}}%

\newcommand{\reffig}[1]{\text{Figure~\ref{#1}}}
\newcommand{\reftab}[1]{\text{Table~\ref{#1}}}
\newcommand{\refeq}[1]{\text{Equation~\ref{#1}}}

\definecolor{demphcolor}{RGB}{144,144,144}
\newcommand{\demph}[1]{\textcolor{demphcolor}{#1}}

\def\paperID{15251} % *** Enter the Paper ID here
\def\confName{CVPR}
\def\confYear{2024}

\title{Do Vision and Language Encoders Represent the World Similarly?}

\author{Mayug Maniparambil\thanks{joint first authors} \hspace{0.5em}\thanks{ML Labs, Dublin City University} \and Raiymbek Akshulakov\footnotemark[1]\hspace{0.5em}\thanks{University of California Berkeley} \and Yasser Abdelaziz Dahou Djilali\thanks{Technological Innovation Institute}\hspace{0.5em}\footnotemark[2] \and Sanath Narayan\footnotemark[4]  \and Mohamed El Amine Seddik\footnotemark[4] \and Karttikeya Mangalam\footnotemark[3] \and Noel E. O'Connor\footnotemark[2]}

\begin{document}
\maketitle
\begin{abstract}

Aligned text-image encoders such as CLIP have become the de-facto model for vision-language tasks. Furthermore, modality-specific encoders achieve impressive performances in their respective domains. This raises a central question: does an alignment exist between uni-modal vision and language encoders since they fundamentally represent the same physical world? Analyzing the latent spaces structure of vision and language models on image-caption benchmarks using the Centered Kernel Alignment (CKA), we find that the representation spaces of unaligned and aligned encoders are semantically similar. In the absence of statistical similarity in aligned encoders like CLIP, we show that a possible matching of unaligned encoders exists without any training. We frame this as a seeded graph-matching problem exploiting the semantic similarity between graphs and propose two methods - a Fast Quadratic Assignment Problem optimization, and a novel localized CKA metric-based matching/retrieval. We demonstrate the effectiveness of this on several downstream tasks including cross-lingual, cross-domain caption matching and image classification. Code available at github.com/mayug/0-shot-llm-vision.

\end{abstract}

\section{Introduction}

% \begin{figure}
%     \centering
%     \includegraphics[trim={0 31.6cm 0 0cm},clip,width=1\linewidth]{images/teaser_v2.png}
%     \includegraphics[trim={0 0cm 0 4cm},clip,width=1\linewidth]{images/teaser_v2.png}
%     % \caption{{The CKA score between different vision encoders and text encoders illustrates that vision encoders trained on big enough data exhibit similar semantic similarity with an unaligned text encoder- all-roberta-large-v1 in comparison to the semantic similarity between CLIP's vision and text encoder (green star). Here the text encoders for all vision encoders is kept constant as all-roberta-large-v1. } We compute the CKA using the representations from the last layer.}
%     \caption{\textbf{CKA vs data scaling}. The CKA score between different vision encoders and text encoders illustrates that vision encoders trained on large enough datasets exhibit similar semantic similarity with an unaligned text encoder, in comparison to the semantic similarity between CLIP's vision and text encoder (represented by the green star). In this analysis, the text encoder used for all vision encoders is kept constant as All-Roberta-large-v1 \cite{allroberta}.}
%     \label{fig:cka_teaser}
% \end{figure}

\begin{figure}[t]
    \centering
    \includegraphics[width=\columnwidth]{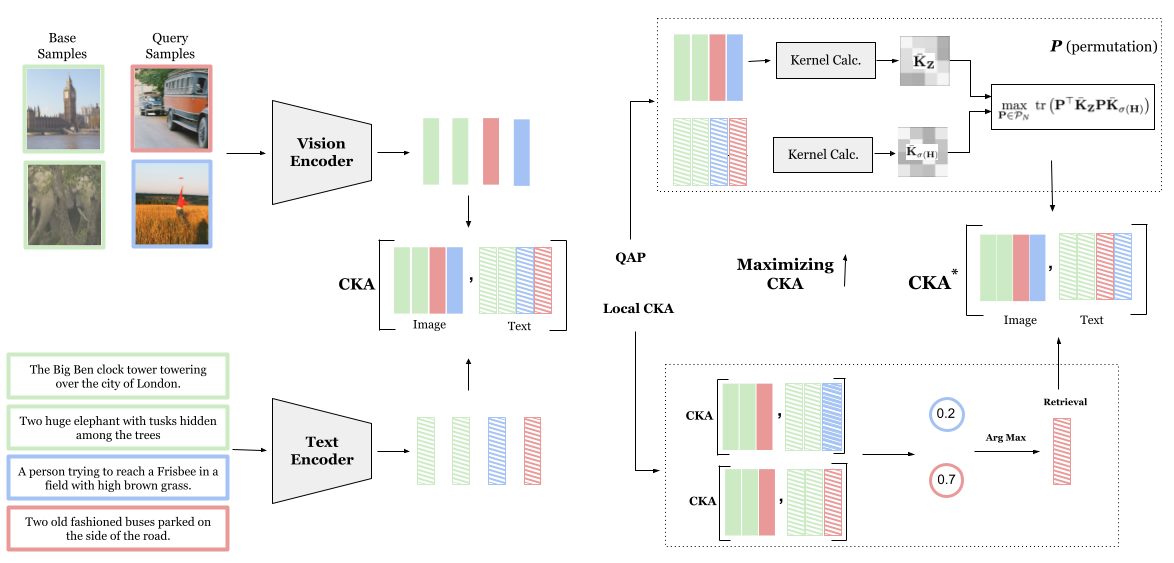}\vspace{-0.2cm}
    % \caption{For retrieval, we append query embeddings to base embeddings and retrieve the best caption that maximizes the local CKA for a query image. For matching, we calculate the kernels for image and text embeddings and employ QAP-based seeded matching to maximize CKA for obtaining the optimal permutation $\bm{P}$. Best viewed zoomed-in.\vspace{-0.50cm}}
        \caption{ For matching, we calculate the kernels for image and text embeddings and employ QAP-based seeded matching to maximize CKA for obtaining the optimal permutation $\bm{P}$. For retrieval, we append query embeddings to base embeddings and retrieve the best caption that maximizes the local CKA for a query image.\vspace{-0.70cm}}
    \label{fig:methods-figure}
\end{figure}

The recent success of deep learning on vision-language tasks mainly relies on jointly trained language and image encoders following the success of CLIP and ALIGN \cite{jia2021scaling, radford2021learning}. The standard procedure for training these models aims at aligning text and image representation using a contrastive loss that maximizes the similarity between image-text pairs while pushing negative captions away \cite{gutmann2010noise, oord2018representation, chen2020simple}. This achieves a statistical similarity across the two latent spaces, which is key to retrieving the closest cross-modal representations using cosine similarity. This property is not valid for unaligned encoders, hence, extra transformations are needed to bridge the gap. These transformations can be training a mapping network that captures the prior distribution over the text and image representations ~\cite{merullo2022linearly, moschella2022relative, norelli2022asif}. The work of \cite{merullo2022linearly} has shown that it is possible to train a linear mapping from the output embeddings of vision encoders to the input embeddings of language models and exhibit impressive performance on image captioning and VQA tasks. This indicates that the representations between the unaligned uni-modal vision and language encoders are sufficiently high level and differ only by a linear transformation. However, this linear layer is trained on CC-3M~\cite{changpinyo2021conceptual} consisting of three million image-caption pairs.

Is this training step necessary? In an ideal scenario, we anticipate an alignment between vision and language encoders as they inherently capture representations of the same physical world. To this end, we employ Centered Kernel Alignment (CKA) ~\cite{raghu2021vision, cortes2012algorithms, kornblith2019similarity}, which is known for measuring representation similarity both within and between networks. As shown in \reffig{fig:bob_qap_cka}, we measure the CKA between a variety of unaligned vision and language encoders ~\cite{dosovitskiy2020image, woo2023convnext, allroberta, oquab2023dinov2, caron2021emerging}, on the image-caption pairs of the COCO~\cite{mscoco} dataset and observe that some have comparable scores to that of aligned encoders like CLIP~\cite{radford2021learning}, affirmative of semantic similarities.

We then ask the question: If the unaligned image and text encoders are semantically similar, is there a way to connect them in a \textit{zero-shot manner?} Do they build a similar representation graph over the same information coming from the two modalities?
We study these questions, revealing key similarities between unaligned image and text encoders, and how these similarities can be exploited for downstream tasks. Furthermore, we devise a caption matching downstream task and show using two novel methods that latent space communication between unaligned encoders could be achieved by leveraging the semantic similarities between the cross-modal spaces. Our contributions are:

\begin{itemize}

    \item We present a matching method that seeks to find the permutation of the captions that maximizes the CKA (see Fig.~\ref{fig:methods-figure}). Hence, We formulate maximizing CKA as a quadratic assignment problem and introduce transformations and normalizations that greatly improve the matching performance.

    \item We propose a local CKA metric and use it to perform retrieval between two unaligned embedding spaces, demonstrating superior performance with that of relative representations~\cite{moschella2022relative} on the COCO caption image retrieval.

    \item The method is benchmarked on COCO, NoCaps~\cite{agrawal2019nocaps} cross-domain caption and image retrieval as well ImageNet-100~\cite{imagenet} classification tasks despite our method not being optimized to align the representation in any manner demonstrating zero-shot communication between the encoder's latent spaces.

    \item Finally, we show a practical application of our method on cross-lingual image retrieval by making use of sentence transformers trained in various languages and a CLIP vision encoder trained only in English. 
    
\end{itemize}
\section{Related Work}

Recently, there has been an increasing consensus that good networks, when trained independently, learn general representations across different architectures and tasks. On the one hand, the works of~\cite{morcos2018insights,li2015convergent,kornblith2019similarity,bonheme2022variational} show that these networks exhibit representation similarity by learning similar latent spaces when trained on similar tasks and data \cite{tsitsulin2019shape,barannikov2021representation,vulic2020all,conneau2017word,lenc2015understanding,mikolov2022exploiting,antonello2021low}. Specifically, \cite{kornblith2019similarity} introduced centered kernel alignment (CKA) as a similarity metric for comparing the inner representations across networks. The CKA measure mitigates the limitation of canonical correlation analysis (CCA) \cite{raghu2017svcca} being invariant to an invertible linear transformation that often leads to difficulty in measuring meaningful similarities between representations. \cite{wu2020similarity} uses CKA for comparing the representations from different layers of different language models and the effect of downstream task-finetuning on the representation similarities, while \cite{bonheme2022variational} utilizes CKA along with Procrustes similarity for understanding the ability of variational autoencoders (VAEs)~\cite{kingma13iclr} in learning disentangled representations. In general, these approaches study the representation similarity in unimodal models, either vision or language. Clearly, however, the use of CKA has been limited to visualization and analysis purposes, whereas we attempt at exploiting CKA as an optimization objective.

Recent works ~\cite{moschella2022relative, norelli2022asif} employ relative representations to match embeddings of unaligned encoders using the cosine similarity to a set of anchors. However, these relative representations are sensitive to the selection of anchors and noise in the original embeddings. Similarly, approaches~\cite{bansal2021revisiting,csiszarik2110similarity} analyze networks and empirically verify the ``good networks learn similar representations" hypothesis by utilizing model stitching~\cite{lenc2015understanding}, which introduces trainable stitching layers to enable swapping parts of different networks. LiMBeR~\cite{merullo2022linearly} can be seen as stitching the output of an image encoder to the input of a language model in the form of soft prompts~\cite{lester2021power}. However, these approaches involve training of stitching layers for evaluating the representation similarity between two models.

In this work, we argue that using an explicit similarity measure as done in ~\cite{moschella2022relative, norelli2022asif}  is sensitive to the selection of anchors and noise in the original embeddings. One design choice is an implicit measure that captures the similarity of similarities, hence, inducing more robustness to the alignment process. Furthermore, we explore how this similarity can be leveraged for downstream cross-modal tasks in a \textit{training-free} manner with the aid of CKA and a set of parallel anchors in the image and text latent embedding spaces. 
\section{Preliminaries}
Centered Kernel Alignment (CKA) has shown its relevance in understanding and comparing the information encoded by different layers of a neural network \cite{kornblith2019similarity}. Formally,  CKA relies on two sets of data $\rmX\in \sR^{p\times N}$ and $\rmY\in \sR^{q\times N}$ through their corresponding kernels $\rmK=k(\rmX^\top, \rmX)\in \sR^{N\times N} $ and $\rmL = \ell(\rmY^\top, \rmY) \in \sR^{N\times N}$ where $k,\ell$ are some kernel functions applied on the columns of $\rmX$ and $\rmY$ respectively (e.g., linear or RBF kernels). Therefore, the CKA is computed in terms of $\rmK$ and $\rmL$ as:
\begin{align}
    \cka(\rmK, \rmL) = \frac{ \hsic(\rmK, \rmL) }{ \sqrt{ \hsic(\rmK, \rmK) \hsic(\rmL, \rmL) } },
\end{align}
where $\hsic(\cdot, \cdot)$ is the Hilbert-Schmidt Independence Criterion \cite{gretton2005measuring, ma2020hsic} defined as:
\begin{align}
    \hsic(\rmK, \rmL) = \frac{1}{(N-1)^2} \tr \left( \rmK \rmC \rmL \rmC \right),
\end{align}
with $\rmC = \rmI - \frac1N \vone \vone^\top$ the centring matrix. We refer the reader to \cite{kornblith2019similarity} for broader properties and studies of the CKA metric on neural network representations.

\section{Proposed Method}

% The goal is to exploit the semantic similarity across unaligned image and text encoders in order to solve vision-language tasks. Given the fact that independently trained vision and language encoders exhibit some inherent semantic similarities in terms of the CKA metric. Hence, we investigate whether maximizing the CKA can be used to bridge the gap between semantic and statistical similarities of unaligned encoders. Precisely, given a set of image representations $\rmX$ and the corresponding ordered captions representations $\rmY$, the obtained CKA score is relatively high as reported in Table \ref{tab:cka_shuffle}. However, we find that the CKA is sensitive to the data ordering since it monotonic decreases with random shuffling. This motivates our methodology for finding an optimal permutation of the data that maximizes the CKA.

Consider a set of $N$ image-caption pairs, $\mathcal{S} = \{ (\vx_{i},\vc_{i} )\}_{i=1}^{N}$, where $\vx_{i} \in \gX $ and $\mathbf{c}_{i} \in \gC $ represent the $i$-th image and its corresponding caption, respectively. In this particular example, we are performing caption-to-image retrieval, but it is applicable for the reverse as well. Let $\vf:\gX\mapsto \sR^{d_1}$ and $\vg:\gC\mapsto \sR^{d_2}$ denote some vision and language encoders respectively. The image-caption pairs are mapped into their corresponding sets of representations $\rmZ = [\vz_1, \ldots, \vz_N] \in \sR^{d_1\times N}  $ and $\rmH = [\vh_1, \ldots, \vh_N] \in \sR^{d_2\times N}$, where $\vz_i = \vf(\vx_i)$ and $\vh_i = \vg(\vc_i)$.

As shown in \reftab{tab:cka_shuffle}, the maximum CKA score is obtained on the ground-truth ordering of the representations $\cka_{\text{max}} = \cka(\rmK_{\rmZ}, \rmK_{\rmH})$, where $\rmK_{\rmZ}$ and $\rmK_{\rmH}$ are the kernels for the image and text representations, defined respectively as $\rmK_\rmZ = k(\rmZ^\top, \rmZ)$ and $\rmK_\rmH = k(\rmH^\top, \rmH)$. We find that the CKA is sensitive to the data ordering. Specifically, we shuffle x\% of data to obtain wrong matches while keeping the remaining 100-x\% aligned, measure the CKA on each new data set, and observe that it monotonically decreases with random shuffling. This motivates our methodology for finding an optimal permutation of the image data that maximizes the CKA.

\begin{table}[t]
    \centering
        \caption{\textbf{CKA reduces with shuffling.} We measure the CKA score between DINOv2~\cite{oquab2023dinov2} and All-Roberta-large-v1~\cite{allroberta} on the 5k COCO~\cite{mscoco} image-caption representations pairs of the valset. The exact ordering yields the best score, whereas randomly shuffling the representations reduces the CKA score. \vspace{-0.2cm}}
    \begin{tabular}{ccccccc}
    \midrule
      Shuffling (\%) &  0  &  20  &  40  &  60  &  80  &  100  \\
    \midrule
           CKA Score        & 0.72  & 0.46  & 0.27  & 0.13  & 0.04  & 0.01  \\
    \midrule
    
    \end{tabular}
    \label{tab:cka_shuffle}
    
\end{table}

Formally, let $\sigma$ be some permutation of the set $\{1, \cdots, N \}$ and denote $\sigma(\rmZ) = [\vz_{\sigma(1)}, \cdots, \vz_{\sigma(N)}]\in \sR^{d_1\times N}$ the set of permuted image representations by $\sigma$. If $\sigma$ is not identity, it disrupts the original ordering of the image representations leading to a lower CKA score as shown in \reftab{tab:cka_shuffle}. Therefore, our goal is to find a permutation $\sigma^*$ that maximizes the CKA. Formally:
\begin{align}\label{eq_optim_sigma}
    \sigma^* = \argmax_{\sigma} \cka(\rmK_{ \sigma(\rmZ) }, \rmK_{\rmH}).
\end{align}
% 
%Here, $\rmK_{\pi_{\mathcal{Z}}(\mathcal{Z})}$ represents the kernel matrix computed after applying the permutation to the image representations, and $\rmL_{\mathcal{H}}$ is the fixed kernel matrix for the text representations. 
The solution to this problem seeks to realign the permuted set of images in a way that maximizes the CKA, potentially recovering the ground-truth pairing between images and their corresponding captions.

To solve the aforementioned optimization problem, we explore two main approaches (visualized in Fig.~\ref{fig:methods-figure}): the Quadratic Assignment Problem (QAP) algorithm and Local CKA-based retrieval and matching. The QAP algorithm provides a global matching solution, seeking the optimal permutation across the query set considered. On the other hand, Local CKA-based retrieval and matching focuses on aligning images and captions using a localized metric, facilitating retrieval on a more granular level. This approach is more suitable where a single query image is given for a set of captions or \textit{vice versa}. %The choice between two methods will depend on the downstream task.

\subsection{QAP Matching}
\label{subsec:qap_matching}
For some random permutation $\sigma$, the optimization problem in \refeq{eq_optim_sigma} can be reformulated as a quadratic optimization problem \cite{vogelstein2015fast} which reads as:
\begin{align}\label{eq_optim_qap}
    \max_{\rmP \in \gP_N } \, \tr \left( \rmP^\top \bar \rmK_{\sigma(\rmZ)} \rmP \bar\rmK_\rmH \right),
\end{align}
where $\gP_N$ is the set of all permutation matrices of size $N$ and $\bar\rmK = \hsic(\rmK, \rmK)^{-\frac12} \rmK \rmC $ stands for the centered and re-scaled kernel. In principle, maximizing the above objective is a relaxation of a graph-matching problem. Moreover, finding a global maximum of \refeq{eq_optim_qap} is NP-hard due to the combinatorial nature of the problem and therefore optimizing it can lead to sub-optimal or approximate solutions.

To overcome the NP-hardness of QAP, in practice, we suppose that we have access to a base set $\gB = \{ (\vz_{i}^b,\vh_{i}^b )\}_{i=1}^M$ of image-caption representations pairs and solve an equivalent objective to \refeq{eq_optim_qap} only partially on some unmatched query set $\gQ = \{\vz_{i}^q \}_{i=1}^N \times \{\vh_{i}^q \}_{i=1}^N$ using a seeded version of the fast QAP algorithm \cite{fishkind2019seeded}. 
Formally, let $\rmZ = [\vz_1^b, \cdots, \vz_M^b, \vz_1^q, \cdots, \vz_N^q]\in \sR^{d_1\times(M+N)}$ and $\rmH = [\vh_1^b, \cdots, \vh_M^b, \vh_1^q, \cdots, \vh_N^q]\in \sR^{d_2\times(M+N)}$ be the matrix concatenating all base and query representations of images and captions respectively, and denote by $\bar \rmK_\rmZ, \bar \rmK_\rmH \in \sR^{(M+N)\times (M+N)}$ the corresponding centered and re-scaled kernels.
The partial matching for aligning the query samples is then performed by solving the following:
\begin{align}
    \max_{\rmP \in \gP_N} \, \tr \left( (\rmI_M \oplus \rmP)^\top \bar \rmK_\rmZ  (\rmI_M \oplus \rmP) \bar \rmK_\rmH \right),
\end{align}
where $\rmI_M \oplus \rmP \in \sR^{(M+N)\times (M+N)} $ stands for the block-diagonal matrix having diagonal blocks $\rmI_M$ and $\rmP$.

%\textbf{Quadratic Assignment algorithm}
%As described subsequently, our method relies on finding the optimal permutation of the data that maximizes the CKA, which can be formulated as a quadratic assignment problem. Specifically, denoting $\sP = \{ \rmP \in  \}$

%stretching and clustering

\subsection{Local CKA based Retrieval and Matching}
\label{subsec:local_cka}

% We extend the concept of a global CKA metric to derive local similarity measures. Analogous to the QAP method, we start with a base set $\gB = \{ (\vz_{i}^b,\vh_{i}^b )\}_{i=1}^M$ of aligned pairs of images and captions representations. The primary objective is to perform image-caption retrieval or matching within an unaligned query set $\gQ = \{\vz_{i}^q \}_{i=1}^N \times \{\vh_{i}^q \}_{i=1}^N$.

% We introduce a local CKA score, denoted as $\localcka(\vz^q, \vh^q)$ for some couple $(\vz^q, \vh^q)\in \mathcal{Q}$ by computing a global CKA score for the image-caption pairs in $\mathcal{B}$, augmented with the query pair $(\vz^q, \vh^q)$. Formally, the local CKA is computed as:

The concept of a global CKA metric is extended to derive local similarity measures suitable for retrieval. This process begins with a base set $\gB = \{ (\vz_{i}^b,\vh_{i}^b )\}_{i=1}^M$ consisting of aligned pairs of images and captions representations. The objective is to facilitate caption-image retrieval/matching within an unaligned query set $\gQ = \{\vz_{i}^q \}_{i=1}^N \times \{\vh_{i}^q \}_{i=1}^N$.

A local CKA score, denoted as $\localcka(\vz^q, \vh^q)$ for a couple $(\vz^q, \vh^q)\in \mathcal{Q}$ is calculated by computing a global CKA score for the image-caption pairs in $\mathcal{B}$, augmented with the query pair $(\vz^q, \vh^q)$. The local CKA is computed as follows:
\begin{equation}
\localcka(\vz^q, \vh^q) = \cka( \rmK_{[\rmZ, \vz^q]} , \rmK_{[\rmH, \vh^q]}),
\end{equation}
where $[\rmM, \vv]$ denotes the concatenation of the matrix $\rmM$ and the vector $\vv$ column-wise and $\rmZ = [\vz_1^b, \cdots, \vz_M^b]\in \sR^{d_1\times M}$ and $\rmH = [\vh_1^b, \cdots, \vh_M^b]\in \sR^{d_2\times M}$.
% 
%where $K_{p}$ represents the kernel computed over $concat(\{x_{i}^b\}_{i=1}^M, \{x_{p}^q\})$, and $K_{g}$ is the kernel for $concat(\{y_{j}^b\}_{j=1}^M, \{y_{g}^q\})$. 
In essence, a correctly matched image-caption pair in $\mathcal{Q}$ would exhibit a higher degree of alignment with the base set $\mathcal{B}$ in terms of the CKA score, resulting in an elevated $\localcka$ score. This metric can be used to calculate a score between one source query and $N$ target queries enabling effective retrieval. Furthermore, this framework allows for the use of linear sum assignment \cite{kuhn1955hungarian} for matching tasks.

\subsection{Stretching and Clustering}
The choice of base samples and the spread of the representations in each embedding space affect the performance of the QAP and Local CKA algorithms. To spread the representations out in each domain for matching, we introduce a stretching matrix that normalizes the features of each dimension by the variance calculated from the query and base sets. Given $\rmX = [\vx_1, \cdots, \vx_d]^\top\in \sR^{d\times N}$, the stretched matrix $\rmX_s$ is computed as 
$\rmX_s = \rmS \rmX$, where the stretching matrix $\rmS \in \sR^{d\times d}$ is a diagonal matrix with inverse empirical standard deviation of the feature dimension as entries, \ie, $\rmS = \text{diag}\left(\frac{1}{\text{std}(\vx_1)}, \cdots, \frac{1}{\text{std}(\vx_d)}\right)$ and $\vx_i \in \sR^N$ is the $i^{th}$ row of $\rmX$. 
% $\rmX_s = \left[ \frac{\vx_1}{ \sqrt{\Var(\vx_1)} }, \cdots, \frac{\vx_N}{ \sqrt{\Var(\vx_N)} } \right] \in \sR^{d\times N} $ where $\Var(\vx_i)$ is the empirical variance computed along the entries of the vector $\vx_i$. 
This stretching operation is performed for both the image and text before calculating the kernels for both QAP and local CKA matching algorithms. For picking the most effective base samples, we find that the simple $k$-means clustering on the image embeddings works best. 
% We do this for both relative representations as well as our methods. 
An ablation on how these affect the QAP and local CKA matching and retrieval accuracies is provided in Sec ~\ref{sec:ablation}.

\section{Experiments}

We assess the performance of the proposed method using various vision and language encoders on a set of downstream tasks. We first detail the encoders, datasets, downstream tasks, and the baselines used. 

\subsection{Vision and Language Encoders}

% We cover vision encoders of different architectures (ViTs \cite{dosoViTskiy2020image} and ConvNeXt \cite{liu2022convnet}) trained in a: Supervised, Language supervised, and Self-supervised way (i.e. contrastive as well as masked image loss), on different training data regimes. For the language encoder, it is required to have an encoder that is capable of producing a global embedding for a caption, as well as having multiple architectures of different sizes, languages, and training data sizes. Thus, using the huggingface's sentence-transformers library where each sentence transformer is first pretrained on the masked language modelling task on a huge text corpus, followed by a finetuning stage on a sentence pairs dataset using a contrastive loss (cite). 

% We report the CKA and Matching Score (MS) of the various combinations of vision and language encoders (see Table. apx). Overall, we find the All-Roberta-large-v1 (cite) shows the best CKA/MS for all vision models, hence, we fix it as the primary choice of the language encoder for all the following analysis and tasks unless otherwise specified.

The experimental setup covers vision encoders of different architectures, such as ViTs \cite{dosovitskiy2020image} and ConvNeXt \cite{liu2022convnet}, trained in various ways: supervised, language-supervised, and self-supervised, across different training data regimes. For the language encoder, an encoder capable of producing a global embedding for a caption is essential. This includes encoders of multiple architectures varying in size, languages, and training data sizes. The Huggingface's sentence-transformers~\cite{sentencetransformer} library is utilized, where each sentence transformer is first pre-trained on the masked language modeling task using a large text corpus, followed by a finetuning stage on a sentence pairs dataset with a contrastive loss. It's not straightforward to acquire a global sentence embedding from decoder-only models like GPT models \cite{radford2019language, brown2020language}, hence we did not study the semantic alignment of these class of models to vision encoders.

The CKA and Matching Score (MS) of the various combinations of vision and language encoders are reported in supplementary. The findings indicate that the All-Roberta-large-v1 \cite{allroberta} demonstrates the best CKA/MS across all vision models, establishing it as the primary language encoder for subsequent tasks, unless specified otherwise.

\subsection{Baselines}
Here, we briefly describe three baselines that we compare our methods against for caption matching/retrieval, image classification, and cross-lingual tasks.

\noindent\textbf{Linear Regression:}
We propose a baseline that learns a linear transformation from the image embedding space to the text using $M$ aligned base examples and apply the transformation to the query image embeddings.  Concretely, given query image embeddings $\rmZ^q = [\vz_1^q, \cdots, \vz_N^q] \in \sR^{d_1\times N}  $ and text embeddings $\rmH^q = [\vh_1^q, \cdots, \vh_N^q] \in \sR^{d_2\times N}$, and a set of aligned base samples $\rmZ^{b} = [\vz_1^{b}, \cdots, \vz_N^{b}] \in \sR^{d_1\times M}  $ and $\rmH^{b} = [\vh_1^{b}, \cdots, \vh_N^{b}] \in \sR^{d_2\times M}$, we first construct a linear transformation between $\rmZ^{b}$ and $\rmH^{b}$ by minimizing the MSE loss as $\rmW = \argmin_\rmW  \Vert \rmW^\top\rmZ^{b} - \rmH^{b} \Vert_F^2 $. Then we use $\rmW$ to transform the query image embeddings $\rmZ^q$ to the text domain as $\hat{\rmH}^q = \rmW^\top \rmZ^q$. Cosine similarity on $\hat{\rmH}^q$ and $\rmH^q$ can be used to perform caption retrieval. 
% For caption matching, we construct a cost matrix $\rmC$ using cosine similarities between $\hat{\rmH}^q$ and $\rmH^q$ and use linear sum assignment to find the permutation matrix.

\noindent\textbf{Relative Representations~\cite{moschella2022relative}:} enable latent space communication between unaligned encoders by representing each query point relative to an aligned base set. Concretely, let $\ell_2$-normalized embeddings for image and text queries be $\rmZ^q = [\vz_1^q, \cdots, \vz_N^q] \in \sR^{d_1\times N}  $ and $\rmH = [\vh_1^q, \cdots, \vh_N^q] \in \sR^{d_2\times N}$, respectively. Utilizing a set of aligned base sample $\ell_2$-normalized embeddings $\rmZ^{b} = [\vz_1^{b}, \cdots, \vz_M^{b}] \in \sR^{d_1\times M}  $ and $\rmH^{b} = [\vh_1^{b}, \cdots, \vh_M^{b}] \in \sR^{d_2\times M}$, we can construct relative image and text query representations as $\rmZ_{rel}^q = (\rmZ^b)^\top \rmZ^q $ and $\rmH_{rel}^q = (\rmH^{b})^\top \rmH^q$.
% \frac{\rmZ^{b\top} \rmZ}{\|Z^{b\top} Z\|_2}
% and $\rmH_{rel} = \frac{H^{b\top} H}{\|H^{b\top} H\|_2}$. 
Relative representations are a single vector of dimension $M$ for each query specifying the cosine similarity of a query sample with all the base samples. Now we can use the cosine similarity on the relative representations to perform retrieval. Sec \textcolor{red}{D} in appendix provides a further comparison with our method.

\noindent\textbf{CLIP~\cite{radford2021learning}:}  
We also compare against CLIP which has been contrastively trained to obtain a joint embedding space- as an upper limit on performance for both retrieval and matching tasks. We perform retrieval using cosine similarity % and matching using a cost matrix constructed from cosine similarity and perform linear sum assignment to find the correct permutation matrix.

For all 3 methods, caption matching can be achieved by constructing a cost matrix using cosine similarities and using linear sum assignment to find the permutation matrix.

\subsection{Downstream Tasks}

% \textbf{Caption Matching.} Given $N$ query images and their corresponding captions, we construct a query set by shuffling the captions. The task is to find the correct permutation over captions for the perfect matching. For \textbf{Retrieval}, we assume access to a single image query and a set of $M$ captions, and the task is to retrieve the correct caption from the overall set. We investigate the alignment between unaligned vision and text encoders using our method on the  MSCOCO and NoCaps validation sets.

% MSCOCO Captions comprises over 120k images, each annotated with multiple human-generated captions. The images are diverse and capture a wide range of scenes, objects, and activities in everyday contexts. This diversity makes the dataset particularly valuable for assessing the quality of the unimodal representations by evaluating the performance on a caption-matching task. Specifically, we test our methods on the validation set of 5k image-caption pairs of the MSCOCO dataset.
\begin{table*}[]
    \centering
    \caption{\textbf{Caption matching and retrieval task performance comparison in cross-domain and in-domain settings.} Base samples from COCO are utilized for matching/retrieval tasks on queries from NoCaps (cross-domain) and COCO (in-domain). CLIP-V denotes the vision encoder of CLIP~\cite{radford2021learning}. We use the Large version of all vision encoders. Table \textcolor{red}{A.5} shows the reverse setting. \vspace{-0.2cm} }
    \begin{tabular}{ll|cc|cc}
    \toprule[0.1em]
    \multirow{2}{*}{\textbf{Method}}  &   \multirow{2}{*}{\textbf{ Vision Model}} & \multicolumn{2}{c|}{\textbf{NoCaps}~\cite{agrawal2019nocaps}} & \multicolumn{2}{c}{\textbf{COCO}~\cite{mscoco}} \\
    & & Matching accuracy &  Top-5 retrieval &  Matching accuracy &  Top-5 retrieval  \\
    \specialrule{1pt}{1pt}{1pt}%
    Cosine Similarity* & CLIP~\cite{radford2021learning} &  \demph{99.5} &  \demph{99.6} & \demph{97.1} & \demph{96.1} \\
    \specialrule{0.5pt}{0.5pt}{0.5pt}%
    \multirow{3}{*}{Linear regression} & CLIP-V~\cite{radford2021learning} & 29.3 & 44.7 & 42.7 &  59.1 \\
             &  ConvNeXt~\cite{woo2023convnext} & 19.0 & 28.5 & 31.3 & 46.1 \\
             &    DINOv2~\cite{oquab2023dinov2} & 38.1 & 50.3 & 45.1 & 65.4 \\
    \specialrule{0.5pt}{0.5pt}{0.5pt}%
    \multirow{2}{*}{Relative}  & CLIP-V~\cite{radford2021learning} & 61.3 & 37.6 & 61.6 &  41.3 \\
      \multirow{2}{*}{representations~\cite{moschella2022relative}} &  ConvNeXt~\cite{woo2023convnext} &  25.5 & 17.8 & 38.6 & 34.1 \\
             &    DINOv2~\cite{oquab2023dinov2} &  46.0 & 46.4 & 47.7 & 52.3 \\
    \specialrule{0.5pt}{0.5pt}{0.5pt}%
    \multirow{3}{*}{\textbf{Ours: QAP}} & CLIP-V~\cite{radford2021learning} & \textbf{67.3} & - & \textbf{72.3} & - \\
             &  ConvNeXt~\cite{woo2023convnext} &  46.7 & - & 66.1 & - \\
             &    DINOv2~\cite{oquab2023dinov2} &  57.7 & - & 66.0 & - \\
    \specialrule{0.5pt}{0.5pt}{0.5pt}%
    \multirow{3}{*}{\textbf{Ours: Local CKA}} & CLIP-V~\cite{radford2021learning} & 65.1 & 60.5	& 71.9 & 69.9 \\
             &  ConvNeXt~\cite{woo2023convnext} &  43.7 & 44.4 & 64.8 &	65.5 \\
             &    DINOv2~\cite{oquab2023dinov2} &  58.7 & \textbf{61.8} & 64.3 &	\textbf{70.5} \\
    \bottomrule[0.1em]
    \end{tabular}
    
    \label{tab:nocap}
\end{table*}

\noindent\textbf{Caption Matching:} Given $N$ query images and their corresponding captions, a query set is constructed by shuffling the captions. The task involves finding the correct permutation over captions for perfect matching. In \textit{Retrieval}, the objective is, given one caption, to retrieve the correct image from the overall set of $N$ images. The alignment between unaligned vision and text encoders is investigated using our methods on the COCO and NoCaps validation sets.

The COCO dataset \cite{mscoco} comprises over 120,000 images with multiple captions per image. It is used for testing unimodal representation quality via a caption-matching task, utilizing a validation set of 5,000 image-caption pairs.
The NoCaps dataset \cite{agrawal2019nocaps} is designed for testing image captioning models on unseen objects, with 166,100 captions for 15,100 images from OpenImages. Its validation set includes novel concepts absent from COCO.

\noindent\textbf{Cross-lingual Caption Matching/Retrieval:}
The task mirrors prior matching and retrieval but uses multilingual captions, say German. Given $N$ images and shuffled German captions, the objective is to match each image with the correct caption. In retrieval, the goal is to select the most fitting German caption for a given query image from the set.

The XTD-10 dataset \cite{aggarwal2020towards} enhances COCO2014 with 1,000 human-annotated multi-lingual captions in ten languages for cross-lingual image retrieval and tagging, serving as a zero-shot model benchmark.

\textbf{ImageNet-100 Classification.}
%  %
The task setup is similar to the conventional classification task with small differences to account for the methods used. Given $N$ query images and their corresponding classes, image representations are obtained by processing them through a vision encoder. In parallel, textual representations are generated in a multi-step process. Initially, several text captions are derived from the class-associated Wordnet synsets' lemmas, definitions, and hypernyms. These captions are then passed through the language encoder and averaged to get the text representations. The classification task is performed by retrieving the closest text representations to each image representation using our local CKA metric. 
We employ the ImageNet-100 dataset. This dataset is a subset of the larger ImageNet dataset, featuring only 100 classes. It includes 130,000 training images, 50,000 validation images, and 100 classes.

%  %
\subsection{Results}

\textbf{Importance of Good Initialization:} For all tasks, we make use of a set of base samples of size $S$ that is kept fixed at 320 samples. The size of the query set is analogously fixed at 500 samples (see Sec \ref{A_sec:ablation_base_query} for more details). These base samples are selected after clustering the image embeddings and choosing one closest sample to each of the $S$ cluster centers. By aligning the initial samples with the diverse cluster centers, we ensure sufficient coverage of the sample space. This enhances the accuracy of the matching process, as the initial alignment closely mirrors the inherent structure and variability within the data. In the case of linear regression, uniform sampling is employed to select the base samples. For relative representations \cite{moschella2022relative}, the same clustering methodology is applied to select base samples, ensuring a fair and consistent comparison between all methods. 

 %This initialization is crucial as it significantly improves the matching quality.
\noindent\textbf{COCO and NoCaps Caption Matching:}
We present the results of cross-domain and in-domain caption matching/retrieval, as detailed in Table \ref{tab:nocap}. We tested each baseline against three different vision models, while employing a consistent language model—specifically, the all-roberta-large-v1. The vision models utilized are OpenAI's CLIP ViT-L/14, the ConvNeXT-Base model (trained on the ImageNet-22k dataset at a resolution of 224x224), and the ViT-L/14 model trained using the DINOv2 method. It is important to note that the first row of the results table features vision and language models both being OpenAI's CLIP ViT-L/14. To effectively analyze cross-domain capabilities, our experiment design involved the use of the COCO validation set as the source of the base set and the NoCaps validation set for querying. Additionally, in-domain results are shown, when using COCO validation for both base and queries. We uniformly sample the query set and average the results over three different seeds.
Although CLIP's cosine similarity metric emerges as the most robust due to the training paradigm inherent in CLIP models, our methods demonstrate commendable performance without necessitating any training. The DINOv2 model, trained solely through self-supervision, demonstrates the formation of semantic concepts independently of language supervision. This is evident in its remarkable top-5 retrieval scores of 70.5\% and 61.8\% on COCO and NoCaps datasets when coupled with an unaligned language encoder through our Local Kernel CKA method. However, the best-performing vision encoder is CLIP's vision encoder which has been trained using language supervision.

\noindent\textbf{ImageNet-100 Classification:}
In Table \ref{tab:imagenet_class}, we detail the performance of our methods on the ImageNet-100 classification task. Mirroring our approach in cross-domain matching and retrieval, we evaluated three different vision models for each method. Notably, the first row of the table highlights the performance using CLIP's embedding cosine similarity. The results are averaged over three different seeds for sampling the query set. A significant observation from this table is the comparatively narrower performance gap between the CLIP's cosine similarity and our methods, as well as the baseline linear regression method, in contrast to the results observed in cross-domain caption matching/retrieval tasks. 

It is interesting that ConvNeXt encoder trained on ImageNet has a classification top1 accuracy improvement of over 14\% compared to CLIP and Dinov2 while on the caption matching task DinoV2 and CLIP perform much better.

\noindent\textbf{Cross-lingual Caption Retrieval:} The results of cross-lingual caption matching/retrieval are presented in Table \ref{tab:cross-lingual} for the 10 languages in the XTD-dataset. OpenAI CLIP's ViT-L vision encoder, trained on English image-caption pairs, and a multilingual sentence transformer paraphrase-multilingual-mpnet-base-v2 were utilized for this task. The accuracy of CLIP's cosine retrieval method exhibits a significant drop when applied to languages other than English. \Eg, CLIP's retrieval at 5 experiences a drop of 30 points when switching from English to other Latin-alphabet languages (Spanish, French, German, and Italian). For non-Latin alphabet languages such as Korean, Chinese, Turkish, \etc, CLIP's performance decreases substantially, collapsing to zero, primarily due to most words resulting in unknown tokens. In contrast, the  QAP and local CKA matching methods demonstrate consistent performance across all languages, including non-Latin languages, attributing to the robustness of a multilingual sentence transformer trained solely on text. On average, QAP surpasses CLIP by 12\% in the caption matching task and also outperforms other baselines like relative representations and linear regression methods. For retrieval at 5, the local CKA-based method exceeds CLIP's performance by over 17\%.

It is possible to push the performance further by using language-specific sentence encoders and we report these results for a few languages in Sec \ref{A_sec:cross_caps_ret} of supplementary. This is a practical application of our method as we can now turn a well-trained English CLIP model's vision encoder into a CLIP model for any low-resource language if a text-only Sentence Transformer trained on that language is available.
\begin{table}[t]
    \centering
    \caption{\textbf{ImageNet-100 classification performance comparison.} We observe a narrow performance gap between the CLIP model and our methods. CLIP-V denotes the vision encoder of CLIP.\vspace{-0.2cm}}
    \begin{tabular}{ll|cc}
\toprule
\textbf{Method}  &    Vision Model &  Top 1 &  Top 5 \\
\specialrule{1pt}{1pt}{1pt}%
 Cosine Similarity* &      CLIP &             \demph{86.1} &             \demph{99.2} \\
\specialrule{0.5pt}{0.5pt}{0.5pt}%
\multirow{3}{*}{Linear Regression} &      CLIP-V &             76.1 &             93.0 \\
         &  ConvNeXt &             84.5 &             95.4 \\
         &    DINOv2 &             73.5 &             92.1 \\
\specialrule{0.5pt}{0.5pt}{0.5pt}%
\multirow{2}{*}{Relative} &      CLIP-V &              8.90 &             30.3 \\
\multirow{2}{*}{representations~\cite{moschella2022relative}} &  ConvNeXt &              7.20 &             15.7 \\
         &    DINOv2 &             49.7 &             75.5 \\
\specialrule{0.5pt}{0.5pt}{0.5pt}%
\multirow{3}{*}{\textbf{Local CKA}} &      CLIP-V &             68.7 &             91.2 \\
         &  ConvNeXt &             83.3 &             95.8 \\
         &    DINOv2 &             67.7 &             88.3 \\
\bottomrule
\end{tabular}
    
    \label{tab:imagenet_class}

    \vspace{-0.3cm}
\end{table}

\begin{table*}
    \centering
    \caption{\textbf{Cross-Lingual caption matching and retrieval performance comparison.} Using QAP and local CKA-based methods we are able to do cross-lingual caption matching/retrieval using CLIP's ViT-L vision encoder and a multi-lingual sentence transformer paraphrase-multilingual-mpnet-base-v2. While CLIP performs well on the Latin languages, it degrades on non-Latin languages. In comparison, our QAP and Local-CKA-based methods perform comparably in Latin languages while outperforming non-Latin languages, highlighting the efficacy of our training-free transfer approach. See Table \textcolor{red}{A.6} and Table \textcolor{red}{A.7} in appendix for additional results.\vspace{-0.2cm}}
\begin{tabular}{llcc|cccc|ccc}
\toprule
 \multirow{2}{*}{\textbf{Language}} & & \multicolumn{2}{c|}{\textbf{Kernel CKA}}   &  \multicolumn{4}{c|}{\textbf{Matching Accuracy}} & \multicolumn{2}{c}{\textbf{Retrieval @ 5}}  \\
  &   &    CLIP               &     Ours        &      CLIP &  Relative\cite{moschella2022relative} &  Linear &  Ours (QAP)  &      CLIP  &  Ours (Local)           \\
\midrule

\multirow{5}{*}{\textbf{Latin}} 
  &  de       &             0.472 &       0.627 &     41.8 &         35.0 &       34.0 &    39.6 &        65.1 &        56.7 \\
  &  en       &             0.567 &       0.646 &     81.5 &         52.5 &       40.9 &    51.6 &        92.5 &        69.0 \\
  &  es       &             0.471 &       0.634 &     50.2 &         37.8 &       31.7 &    41.4 &        68.5 &        61.6 \\
  &  fr       &             0.477 &       0.624 &     49.4 &         37.5 &       30.7 &    40.2 &        68.7 &        57.6 \\
  &  it       &             0.472 &       0.638 &     41.0 &         37.2 &       34.9 &    38.5 &        61.3 &        59.7 \\

\midrule
 \multirow{6}{*}{\textbf{Non-Latin}} 

 &   jp       &             0.337 &       0.598 &     13.2 &         28.3 &       23.5 &    30.5 &       30.0 &        49.4 \\
  &  ko       &             0.154 &       0.620 &     0.50 &         30.4 &       23.5 &    30.9 &        3.30 &        53.4 \\
   & pl       &             0.261 &       0.642 &     5.40 &         36.6 &       30.2 &    40.2 &        18.8 &        59.5 \\
   & ru       &             0.077 &       0.632 &     0.80 &         31.9 &       30.7 &    35.1 &        4.10 &        53.2 \\
   & tr       &             0.301 &       0.624 &     4.30 &         35.8 &       29.6 &    38.9 &        15.2 &        59.3 \\
   & zh       &             0.133 &       0.641 &     2.70 &         36.5 &       31.1 &    40.3 &        8.90 &        57.8 \\

   \midrule

   & \mycc \textbf{Avg.}       &     \mycc    --     & \mycc  --     &  \mycc 26.4    & \mycc  36.3      &  \mycc 30.9  & \mycc \textbf{38.8} &  \mycc 39.6 &  \mycc \textbf{57.9} \\
\bottomrule
\end{tabular}

    \label{tab:cross-lingual}
\end{table*}

\begin{figure*}
    \centering
    \includegraphics[width=1\linewidth]{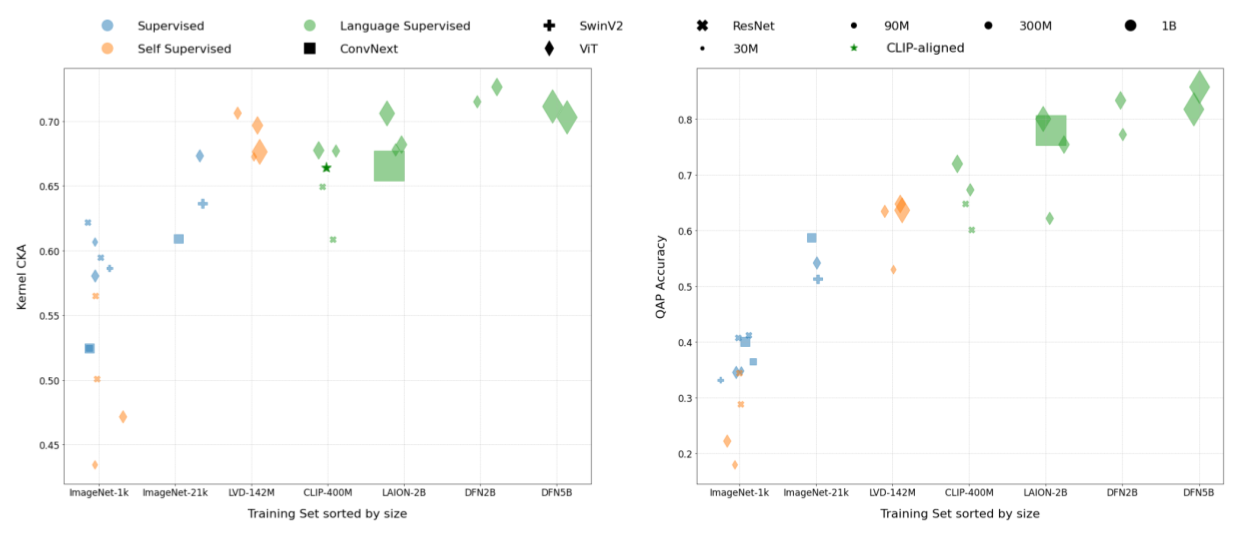}
    \caption{\textbf{Kernel CKA and QAP Matching accuracy are correlated with the training set size and quality of the training set.} Here the language encoder is kept constant to the best BERT-sentence encoder (i.e.All-Roberta-large-v1). There is a clear correlation between CKA and QAP Matching accuracy across all architectures, training paradigm and data regimes.}
    \label{fig:bob_qap_cka}
\end{figure*}

\subsection{Matching complexity}

\begin{table}[]
    \centering
    \caption{Run times for different methods}
    \resizebox{1\columnwidth}{!}{%
    \begin{tabular}{l|llll}
    Method     & QAP        & Local CKA & Relative & Linear   \\
    %\specialrule{1pt}{1pt}{1pt}%
    \toprule[0.07em]
    Run times  & 40 seconds & 5 mins           & 1 second & 1 second \\
    Complexity & $\gO(n^3)$      & $\gO(n^4)$            & $\gO(n^2)$    & $\gO(n\times d)$  \\
     \bottomrule[0.07em]
    \end{tabular}%
    
    }
    \label{tab:run_times}
    \vspace{-0.9cm}
\end{table}

In ~\reftab{tab:run_times}, we go over the time complexity and runtimes of QAP matching and local CKA based retrieval in comparison to the other baselines for matching when number of base samples and query samples are 320, 500 respectively. For all time complexities, we assume number of base samples m to be of the order of the number of query samples n. QAP uses the seeded version of the fast QAP algorithm from the SciPy library, which has a worst time complexity of $\mathcal{O}(n^3)$ \cite{fishkind2019seeded}, while local CKA retrieval requires constructing a graph over all the query image and text pairs, $\mathcal{O}(n^2)$, using local CKA, which is also $\mathcal{O}(n^2)$ resulting in $\mathcal{O}(n^4)$. Relative involves the calculation of the relative representations for every query image and text pair, resulting in a time complexity of $\mathcal{O}(n^2)$, but it's fast due to highly optimized algorithms for matrix multiplications in PyTorch~\cite{paszke2019pytorch}. Linear has a time complexity of $\mathcal{O}(nd)$, where $n$ is the number of samples and $d$ is the number of dimensions. It is to be noted that QAP runs on the CPU, and a CUDA-optimized version could bring the runtimes further down from 40 seconds. An efficient implementation of Local Kernel CKA is also possible, where the CKA of base samples is precalculated, and the graph is constructed in an additive manner, which would bring down the time complexity to $\mathcal{O}(n^3)$. For both relative and linear matching, we make use of SciPy's modified Jonker-Volgenant algorithm \cite{crouse2016implementing} for linear sum assignment, which has the worst time complexity of $\mathcal{O}(n^3)$.

\section{Analysis}
\label{sec:analysis}

This section focuses on how training paradigms, data regimes, and encoder size/architecture influence a vision encoder's ability to represent the world similarly to a language encoder. This is assessed by comparing the semantic alignment of their representation spaces using CKA as well as QAP matching accuracy. \reffig{fig:bob_qap_cka} compares the kernel CKA and caption matching accuracy of different vision encoders with a fixed text-encoder (\ie, All-Roberta-large-v1), against the training datasets on which the vision encoder was trained for all pairs in the COCO captions validation set. The findings are summarized below:

\noindent\textbf{Scale and quality of dataset results in encoders with high semantic alignment with the language space:} It is observed that SSL methods like DINOv2 can learn semantic concepts in a relative manner even without language supervision during training. The CKA and QAP matching accuracy for DINOv2 embeddings are comparable to CLIP models, despite lacking language supervision and having significantly less data (LVD-142's 142M vs Open-AI-CLIP's 400M). A general trend emerges where more training data leads to semantically richer visual embeddings, evident when comparing CKA and QAP Accuracies from ImageNet1K to DFN-5B datasets. Notably, training on a curated dataset proves more effective than on an uncurated dataset of the same size, especially for smaller models. This is illustrated by the higher CKA and QAP accuracy of ViT-Large trained on the curated DFN-2B dataset compared to ViT-Large/Giant, and ConvNext-xxLarge trained on Laion 2B. Additionally, SSL methods show less semantic consistency when trained on ImageNet1K, as indicated by the clear difference in QAP accuracies between DINO trained on ImageNet1K and DINOv2 trained on LVD-142M.\\
\textbf{Vision Encoders Trained with Language Supervision Exhibit Greater Semantic Alignment with Language Encoders:}
In line with the findings of Merullo et al.\cite{merullo2022linearly}, it is observed in our experiments that vision encoders trained with more language supervision on datasets of comparable size exhibit a higher degree of semantic alignment with language encoders compared to self-supervised methods. For example, ViT-Large trained on CLIP-400M with language supervision demonstrates superior caption-matching capabilities compared to DINOv2's ViT-Large trained on LVD-142M. Similarly, we verify that class label supervision, like that from ImageNet, leads to more semantically aligned image encoders when compared to self-supervision when similarly sized models are compared on ImageNet-1k. For example, all supervised encoders trained on ImageNet-1k have higher CKA as well as QAP matching accuracy than all the self-supervised models.

\section{Ablations}
\label{sec:ablation}

This section rationalizes our method choices through ablation studies on clustering, stretching, and the global CKA metric. We demonstrate the impact of these components on the performance of our methods, primarily through \reftab{tab:stretch_ablation}, which delineates the effectiveness of the QAP and the local CKA metric under various configurations. It shows the performance metrics in scenarios where each main component is either integrated or omitted. Notably, in instances where the CKA metric is not used, we opt for normalized correlation matrices for each graph. The empirical results presented are derived from the caption matching/retrieval task, utilizing both base and query sets extracted from the COCO validation set of size 320 and 500 respectively.

\noindent\textbf{Choice of the metric:} CKA is more beneficial than using just the scaled correlation matrix to represent the semantic relationships in an embedding space as matching accuracy increases from 10.1\% to 48.8\%. The choice of a robust metric is core to aligning vision and language latent spaces.

\noindent\textbf{Impact of Stretching:} It is clear that stretching facilitates better alignment of embeddings in our methods as stretching spreads the representations out in each modality without sacrificing the relative positions of the different embeddings within each embedding space. This is reflected in the increase of QAP accuracy from 48.8\% to 57.3\%.

\noindent\textbf{Clustering \emph{vs.} Uniform Sampling:} The choice of the base set is important in QAP matching and local CKA retrieval, as it measures any query pair alignment with the base set. A diverse base set is essential to capture a broad semantic range, and clustering within one of the embedding spaces aids in achieving this diversity. The third and fifth rows of the table demonstrate that clustering enhances the QAP performance from 57.3\% to 65.5\%.
Consequently, these results highlight that all the components together significantly enhance the efficacy of our proposed approach.

\begin{table}
    \centering
    \caption{\textbf{Impact of clustering and stretching.} The matching and retrieval performance is the best when both clustering and stretching are employed. Hence, justifying this choice.\vspace{-0.2cm}}
    \resizebox{1.0\columnwidth}{!}{%
    % \begin{tabular}{ll|rrrr}
    % \toprule
    % & &  QAP &  Local CKA &  Local CKA &  Local CKA \\
    % & &  matching & matching & Top 1 retrieval &  Top 5 retrieval \\
    % \textbf{Stretching} & \textbf{Clustering} & & & &\\
    % \midrule
    % \xmark & \xmark &              49.8 &           49.5 &                  25.6 &                  54.0 \\
    % \cmark & \xmark &              53.1 &           51.2 &                  28.1 &                  56.1 \\
    % \xmark & \cmark &              53.7 &           53.9 &                  29.1 &                  60.0 \\
    % \cmark & \cmark &              57.7 &           55.7 &                  30.3 &                  61.0 \\
    % \bottomrule
    % \end{tabular}%
% \begin{tabular}{lll|ccc}
% \centering
%            &            &      & QAP & Local CKA & Local CKA \\
% Clustering & Stretching & CKA & matching & matching & retrieval5\\
% \specialrule{1pt}{1pt}{1pt}%
% \xmark & \xmark & \xmark  & 10.1         & 16.2               & 1.0                  \\
% \xmark & \xmark & \cmark  & 48.8         & 48.5               & 60.2                 \\
% \xmark & \cmark & \cmark  & 57.3         & 56.7               & 73.0                 \\
% \cmark & \xmark & \cmark  & 56.2         & 55.1               & 66.4                 \\
% \cmark & \cmark & \cmark  & \textbf{65.5} & \textbf{63.3}     & \textbf{77.2}                

% \end{tabular}

\centering
\begin{tabular}{ccc|ccc}
% Column names split into two rows
\multirow{2}{*}{Clustering} & \multirow{2}{*}{Stretching} & \multirow{2}{*}{CKA} & QAP & Local CKA & Local CKA \\
& & & Matching & Matching & Retrieval @ 5 \\

% Horizontal line
%\specialrule{1pt}{1pt}{1pt}
\toprule[0.07em]
% Rows with centered columns
\xmark & \xmark & \xmark  & 10.1 & 16.2 & 1.0 \\
\xmark & \xmark & \cmark  & 48.8 & 48.5 & 60.2 \\
\xmark & \cmark & \cmark  & 57.3 & 56.7 & 73.0 \\
\cmark & \xmark & \cmark  & 56.2 & 55.1 & 66.4 \\
\cmark & \cmark & \cmark  & \textbf{65.5} & \textbf{63.3} & \textbf{77.2} \\
    \bottomrule[0.07em]
    \vspace{-0.6cm}
\end{tabular}

    }
    
    \label{tab:stretch_ablation}
\end{table}

\section{Conclusion}
In this work, we ask the question, \textit{`Do vision encoders and language encoders represent the world similarly?'} and study this using CKA and a caption-matching task. We find that well-trained vision encoders on sufficiently large datasets exhibit surprisingly high semantic similarity with language encoders comparable to aligned encoders, irrespective of the training paradigm. Inspired by this, we draw parallels between CKA and the QAP matching objective and use seeded graph matching to align vision and language encoders by maximizing CKA. We also devise a local CKA-based metric to enable retrieval between unaligned vision and language encoders demonstrating a better performance than that of relative representations on cross-domain and cross-lingual caption matching/retrieval tasks, facilitating zero-shot latent space communication between unaligned encoders. 

% Our analysis demonstrates that semantic similarity exists between unaligned encoders and is likely to influence future research to incorporate pure image encoders with LLMs for better spatial reasoning.
\clearpage  %% Dont remove this. 
{
    \small
    \bibliographystyle{ieeenat_fullname}
    \bibliography{main}

\begin{thebibliography}{48}
\providecommand{\natexlab}[1]{#1}
\providecommand{\url}[1]{\texttt{#1}}
\expandafter\ifx\csname urlstyle\endcsname\relax
  \providecommand{\doi}[1]{doi: #1}\else
  \providecommand{\doi}{doi: \begingroup \urlstyle{rm}\Url}\fi

\bibitem[Aggarwal and Kale(2020)]{aggarwal2020towards}
Pranav Aggarwal and Ajinkya Kale.
\newblock Towards zero-shot cross-lingual image retrieval.
\newblock \emph{arXiv preprint arXiv:2012.05107}, 2020.

\bibitem[Agrawal et~al.(2019)Agrawal, Desai, Wang, Chen, Jain, Johnson, Batra, Parikh, Lee, and Anderson]{agrawal2019nocaps}
Harsh Agrawal, Karan Desai, Yufei Wang, Xinlei Chen, Rishabh Jain, Mark Johnson, Dhruv Batra, Devi Parikh, Stefan Lee, and Peter Anderson.
\newblock Nocaps: Novel object captioning at scale.
\newblock In \emph{Proceedings of the IEEE/CVF international conference on computer vision}, pages 8948--8957, 2019.

\bibitem[Antonello et~al.(2021)Antonello, Turek, Vo, and Huth]{antonello2021low}
Richard Antonello, Javier~S Turek, Vy Vo, and Alexander Huth.
\newblock Low-dimensional structure in the space of language representations is reflected in brain responses.
\newblock \emph{Advances in neural information processing systems}, 2021.

\bibitem[Bansal et~al.(2021)Bansal, Nakkiran, and Barak]{bansal2021revisiting}
Yamini Bansal, Preetum Nakkiran, and Boaz Barak.
\newblock Revisiting model stitching to compare neural representations.
\newblock \emph{Advances in neural information processing systems}, 2021.

\bibitem[Barannikov et~al.(2021)Barannikov, Trofimov, Balabin, and Burnaev]{barannikov2021representation}
Serguei Barannikov, Ilya Trofimov, Nikita Balabin, and Evgeny Burnaev.
\newblock Representation topology divergence: A method for comparing neural network representations.
\newblock \emph{arXiv preprint arXiv:2201.00058}, 2021.

\bibitem[Bonheme and Grzes(2022)]{bonheme2022variational}
Lisa Bonheme and Marek Grzes.
\newblock How do variational autoencoders learn? insights from representational similarity.
\newblock \emph{arXiv preprint arXiv:2205.08399}, 2022.

\bibitem[Brown et~al.(2020)Brown, Mann, Ryder, Subbiah, Kaplan, Dhariwal, Neelakantan, Shyam, Sastry, Askell, et~al.]{brown2020language}
Tom Brown, Benjamin Mann, Nick Ryder, Melanie Subbiah, Jared~D Kaplan, Prafulla Dhariwal, Arvind Neelakantan, Pranav Shyam, Girish Sastry, Amanda Askell, et~al.
\newblock Language models are few-shot learners.
\newblock \emph{Advances in neural information processing systems}, 33:\penalty0 1877--1901, 2020.

\bibitem[Caron et~al.(2021)Caron, Touvron, Misra, J{\'e}gou, Mairal, Bojanowski, and Joulin]{caron2021emerging}
Mathilde Caron, Hugo Touvron, Ishan Misra, Herv{\'e} J{\'e}gou, Julien Mairal, Piotr Bojanowski, and Armand Joulin.
\newblock Emerging properties in self-supervised vision transformers.
\newblock In \emph{Proceedings of the IEEE/CVF international conference on computer vision}, pages 9650--9660, 2021.

\bibitem[Changpinyo et~al.(2021)Changpinyo, Sharma, Ding, and Soricut]{changpinyo2021conceptual}
Soravit Changpinyo, Piyush Sharma, Nan Ding, and Radu Soricut.
\newblock Conceptual 12m: Pushing web-scale image-text pre-training to recognize long-tail visual concepts.
\newblock In \emph{Proceedings of the IEEE/CVF Conference on Computer Vision and Pattern Recognition}, pages 3558--3568, 2021.

\bibitem[Chen et~al.(2020)Chen, Kornblith, Norouzi, and Hinton]{chen2020simple}
Ting Chen, Simon Kornblith, Mohammad Norouzi, and Geoffrey Hinton.
\newblock A simple framework for contrastive learning of visual representations.
\newblock In \emph{International conference on machine learning}, pages 1597--1607. PMLR, 2020.

\bibitem[Conneau et~al.(2017)Conneau, Lample, Ranzato, Denoyer, and J{\'e}gou]{conneau2017word}
Alexis Conneau, Guillaume Lample, Marc'Aurelio Ranzato, Ludovic Denoyer, and Herv{\'e} J{\'e}gou.
\newblock Word translation without parallel data.
\newblock \emph{arXiv preprint arXiv:1710.04087}, 2017.

\bibitem[Cortes et~al.(2012)Cortes, Mohri, and Rostamizadeh]{cortes2012algorithms}
Corinna Cortes, Mehryar Mohri, and Afshin Rostamizadeh.
\newblock Algorithms for learning kernels based on centered alignment.
\newblock \emph{The Journal of Machine Learning Research}, 13\penalty0 (1):\penalty0 795--828, 2012.

\bibitem[Crouse(2016)]{crouse2016implementing}
David~F Crouse.
\newblock On implementing 2d rectangular assignment algorithms.
\newblock \emph{IEEE Transactions on Aerospace and Electronic Systems}, 52\penalty0 (4):\penalty0 1679--1696, 2016.

\bibitem[Csisz{\'a}rik et~al.(2021)Csisz{\'a}rik, Kor{\"o}si-Szab{\'o}, Matszangosz, Papp, and Varga]{csiszarik2110similarity}
Adri{\'a}n Csisz{\'a}rik, P{\'e}ter Kor{\"o}si-Szab{\'o}, Akos~K Matszangosz, Gergely Papp, and D{\'a}niel Varga.
\newblock Similarity and matching of neural network representations.
\newblock \emph{arXiv preprint arXiv:2110.14633}, 2021.

\bibitem[Deng et~al.(2009)Deng, Dong, Socher, Li, Li, and Fei-Fei]{imagenet}
J. Deng, W. Dong, R. Socher, L. Li, K. Li, and L. Fei-Fei.
\newblock Imagenet: A large-scale hierarchical image database.
\newblock In \emph{Proceedings of the IEEE/CVF Conference on Computer Vision and Pattern Recognition}, 2009.

\bibitem[Dosovitskiy et~al.(2020)Dosovitskiy, Beyer, Kolesnikov, Weissenborn, Zhai, Unterthiner, Dehghani, Minderer, Heigold, Gelly, et~al.]{dosovitskiy2020image}
Alexey Dosovitskiy, Lucas Beyer, Alexander Kolesnikov, Dirk Weissenborn, Xiaohua Zhai, Thomas Unterthiner, Mostafa Dehghani, Matthias Minderer, Georg Heigold, Sylvain Gelly, et~al.
\newblock An image is worth 16x16 words: Transformers for image recognition at scale.
\newblock \emph{arXiv preprint arXiv:2010.11929}, 2020.

\bibitem[Fishkind et~al.(2019)Fishkind, Adali, Patsolic, Meng, Singh, Lyzinski, and Priebe]{fishkind2019seeded}
Donniell~E Fishkind, Sancar Adali, Heather~G Patsolic, Lingyao Meng, Digvijay Singh, Vince Lyzinski, and Carey~E Priebe.
\newblock Seeded graph matching.
\newblock \emph{Pattern recognition}, 87:\penalty0 203--215, 2019.

\bibitem[Gretton et~al.(2005)Gretton, Bousquet, Smola, and Sch{\"o}lkopf]{gretton2005measuring}
Arthur Gretton, Olivier Bousquet, Alex Smola, and Bernhard Sch{\"o}lkopf.
\newblock Measuring statistical dependence with hilbert-schmidt norms.
\newblock In \emph{International conference on algorithmic learning theory}, pages 63--77. Springer, 2005.

\bibitem[Gutmann and Hyv{\"a}rinen(2010)]{gutmann2010noise}
Michael Gutmann and Aapo Hyv{\"a}rinen.
\newblock Noise-contrastive estimation: A new estimation principle for unnormalized statistical models.
\newblock In \emph{Proceedings of the Thirteenth International Conference on Artificial Intelligence and Statistics}, pages 297--304. JMLR Workshop and Conference Proceedings, 2010.

\bibitem[Jia et~al.(2021)Jia, Yang, Xia, Chen, Parekh, Pham, Le, Sung, Li, and Duerig]{jia2021scaling}
Chao Jia, Yinfei Yang, Ye Xia, Yi-Ting Chen, Zarana Parekh, Hieu Pham, Quoc Le, Yun-Hsuan Sung, Zhen Li, and Tom Duerig.
\newblock Scaling up visual and vision-language representation learning with noisy text supervision.
\newblock In \emph{International conference on machine learning}, pages 4904--4916. PMLR, 2021.

\bibitem[Kingma and Welling(2014)]{kingma13iclr}
Diederik~P Kingma and Max Welling.
\newblock Auto-encoding variational bayes.
\newblock \emph{ICLR}, 2014.

\bibitem[Kornblith et~al.(2019)Kornblith, Norouzi, Lee, and Hinton]{kornblith2019similarity}
Simon Kornblith, Mohammad Norouzi, Honglak Lee, and Geoffrey Hinton.
\newblock Similarity of neural network representations revisited.
\newblock In \emph{International conference on machine learning}, pages 3519--3529. PMLR, 2019.

\bibitem[Kuhn(1955)]{kuhn1955hungarian}
Harold~W Kuhn.
\newblock The hungarian method for the assignment problem.
\newblock \emph{Naval research logistics quarterly}, 2\penalty0 (1-2):\penalty0 83--97, 1955.

\bibitem[Lenc and Vedaldi(2015)]{lenc2015understanding}
Karel Lenc and Andrea Vedaldi.
\newblock Understanding image representations by measuring their equivariance and equivalence.
\newblock In \emph{Proceedings of the IEEE conference on computer vision and pattern recognition}, pages 991--999, 2015.

\bibitem[Lester et~al.(2021)Lester, Al-Rfou, and Constant]{lester2021power}
Brian Lester, Rami Al-Rfou, and Noah Constant.
\newblock The power of scale for parameter-efficient prompt tuning.
\newblock \emph{arXiv preprint arXiv:2104.08691}, 2021.

\bibitem[Li et~al.(2015)Li, Yosinski, Clune, Lipson, and Hopcroft]{li2015convergent}
Yixuan Li, Jason Yosinski, Jeff Clune, Hod Lipson, and John Hopcroft.
\newblock Convergent learning: Do different neural networks learn the same representations?
\newblock \emph{arXiv preprint arXiv:1511.07543}, 2015.

\bibitem[Lin et~al.(2014)Lin, Maire, Belongie, Hays, Perona, Ramanan, Doll{\'a}r, and Zitnick]{mscoco}
Tsung-Yi Lin, Michael Maire, Serge Belongie, James Hays, Pietro Perona, Deva Ramanan, Piotr Doll{\'a}r, and C~Lawrence Zitnick.
\newblock Microsoft coco: Common objects in context.
\newblock In \emph{ECCV}, 2014.

\bibitem[Liu et~al.(2019)Liu, Ott, Goyal, Du, Joshi, Chen, Levy, Lewis, Zettlemoyer, and Stoyanov]{allroberta}
Yinhan Liu, Myle Ott, Naman Goyal, Jingfei Du, Mandar Joshi, Danqi Chen, Omer Levy, Mike Lewis, Luke Zettlemoyer, and Veselin Stoyanov.
\newblock Roberta: A robustly optimized bert pretraining approach.
\newblock \emph{arXiv preprint arXiv:1907.11692}, 2019.

\bibitem[Liu et~al.(2022)Liu, Mao, Wu, Feichtenhofer, Darrell, and Xie]{liu2022convnet}
Zhuang Liu, Hanzi Mao, Chao-Yuan Wu, Christoph Feichtenhofer, Trevor Darrell, and Saining Xie.
\newblock A convnet for the 2020s.
\newblock In \emph{Proceedings of the IEEE/CVF conference on computer vision and pattern recognition}, pages 11976--11986, 2022.

\bibitem[Ma et~al.(2020)Ma, Lewis, and Kleijn]{ma2020hsic}
Wan-Duo~Kurt Ma, JP Lewis, and W~Bastiaan Kleijn.
\newblock The hsic bottleneck: Deep learning without back-propagation.
\newblock In \emph{Proceedings of the AAAI conference on artificial intelligence}, pages 5085--5092, 2020.

\bibitem[Merullo et~al.(2022)Merullo, Castricato, Eickhoff, and Pavlick]{merullo2022linearly}
Jack Merullo, Louis Castricato, Carsten Eickhoff, and Ellie Pavlick.
\newblock Linearly mapping from image to text space.
\newblock \emph{arXiv preprint arXiv:2209.15162}, 2022.

\bibitem[Mikolov et~al.(2022)Mikolov, Le, and Sutskever]{mikolov2022exploiting}
Tomas Mikolov, Quoc~V Le, and Ilya Sutskever.
\newblock Exploiting similarities among languages for machine translation (2013).
\newblock \emph{arXiv preprint arXiv:1309.4168}, 2022.

\bibitem[Morcos et~al.(2018)Morcos, Raghu, and Bengio]{morcos2018insights}
Ari Morcos, Maithra Raghu, and Samy Bengio.
\newblock Insights on representational similarity in neural networks with canonical correlation.
\newblock \emph{Advances in neural information processing systems}, 31, 2018.

\bibitem[Moschella et~al.(2022)Moschella, Maiorca, Fumero, Norelli, Locatello, and Rodol{\`a}]{moschella2022relative}
Luca Moschella, Valentino Maiorca, Marco Fumero, Antonio Norelli, Francesco Locatello, and Emanuele Rodol{\`a}.
\newblock Relative representations enable zero-shot latent space communication.
\newblock In \emph{The Eleventh International Conference on Learning Representations}, 2022.

\bibitem[Norelli et~al.(2022)Norelli, Fumero, Maiorca, Moschella, Rodola, and Locatello]{norelli2022asif}
Antonio Norelli, Marco Fumero, Valentino Maiorca, Luca Moschella, Emanuele Rodola, and Francesco Locatello.
\newblock Asif: Coupled data turns unimodal models to multimodal without training.
\newblock \emph{arXiv preprint arXiv:2210.01738}, 2022.

\bibitem[Oord et~al.(2018)Oord, Li, and Vinyals]{oord2018representation}
Aaron van~den Oord, Yazhe Li, and Oriol Vinyals.
\newblock Representation learning with contrastive predictive coding.
\newblock \emph{arXiv preprint arXiv:1807.03748}, 2018.

\bibitem[Oquab et~al.(2023)Oquab, Darcet, Moutakanni, Vo, Szafraniec, Khalidov, Fernandez, Haziza, Massa, El-Nouby, et~al.]{oquab2023dinov2}
Maxime Oquab, Timoth{\'e}e Darcet, Th{\'e}o Moutakanni, Huy Vo, Marc Szafraniec, Vasil Khalidov, Pierre Fernandez, Daniel Haziza, Francisco Massa, Alaaeldin El-Nouby, et~al.
\newblock Dinov2: Learning robust visual features without supervision.
\newblock \emph{arXiv preprint arXiv:2304.07193}, 2023.

\bibitem[Paszke et~al.(2019)Paszke, Gross, Massa, Lerer, Bradbury, Chanan, Killeen, Lin, Gimelshein, Antiga, et~al.]{paszke2019pytorch}
Adam Paszke, Sam Gross, Francisco Massa, Adam Lerer, James Bradbury, Gregory Chanan, Trevor Killeen, Zeming Lin, Natalia Gimelshein, Luca Antiga, et~al.
\newblock Pytorch: An imperative style, high-performance deep learning library.
\newblock \emph{Advances in neural information processing systems}, 32, 2019.

\bibitem[Radford et~al.(2019)Radford, Wu, Child, Luan, Amodei, Sutskever, et~al.]{radford2019language}
Alec Radford, Jeffrey Wu, Rewon Child, David Luan, Dario Amodei, Ilya Sutskever, et~al.
\newblock Language models are unsupervised multitask learners.
\newblock \emph{OpenAI blog}, 1\penalty0 (8):\penalty0 9, 2019.

\bibitem[Radford et~al.(2021)Radford, Kim, Hallacy, Ramesh, Goh, Agarwal, Sastry, Askell, Mishkin, Clark, et~al.]{radford2021learning}
Alec Radford, Jong~Wook Kim, Chris Hallacy, Aditya Ramesh, Gabriel Goh, Sandhini Agarwal, Girish Sastry, Amanda Askell, Pamela Mishkin, Jack Clark, et~al.
\newblock Learning transferable visual models from natural language supervision.
\newblock In \emph{International conference on machine learning}, pages 8748--8763. PMLR, 2021.

\bibitem[Raghu et~al.(2017)Raghu, Gilmer, Yosinski, and Sohl-Dickstein]{raghu2017svcca}
Maithra Raghu, Justin Gilmer, Jason Yosinski, and Jascha Sohl-Dickstein.
\newblock Svcca: Singular vector canonical correlation analysis for deep learning dynamics and interpretability.
\newblock \emph{Advances in neural information processing systems}, 30, 2017.

\bibitem[Raghu et~al.(2021)Raghu, Unterthiner, Kornblith, Zhang, and Dosovitskiy]{raghu2021vision}
Maithra Raghu, Thomas Unterthiner, Simon Kornblith, Chiyuan Zhang, and Alexey Dosovitskiy.
\newblock Do vision transformers see like convolutional neural networks?
\newblock \emph{Advances in Neural Information Processing Systems}, 34:\penalty0 12116--12128, 2021.

\bibitem[Reimers and Gurevych(2019)]{sentencetransformer}
Nils Reimers and Iryna Gurevych.
\newblock Sentence-bert: Sentence embeddings using siamese bert-networks.
\newblock In \emph{Proceedings of the 2019 Conference on Empirical Methods in Natural Language Processing}. Association for Computational Linguistics, 2019.

\bibitem[Tsitsulin et~al.(2019)Tsitsulin, Munkhoeva, Mottin, Karras, Bronstein, Oseledets, and M{\"u}ller]{tsitsulin2019shape}
Anton Tsitsulin, Marina Munkhoeva, Davide Mottin, Panagiotis Karras, Alex Bronstein, Ivan Oseledets, and Emmanuel M{\"u}ller.
\newblock The shape of data: Intrinsic distance for data distributions.
\newblock \emph{arXiv preprint arXiv:1905.11141}, 2019.

\bibitem[Vogelstein et~al.(2015)Vogelstein, Conroy, Lyzinski, Podrazik, Kratzer, Harley, Fishkind, Vogelstein, and Priebe]{vogelstein2015fast}
Joshua~T Vogelstein, John~M Conroy, Vince Lyzinski, Louis~J Podrazik, Steven~G Kratzer, Eric~T Harley, Donniell~E Fishkind, R~Jacob Vogelstein, and Carey~E Priebe.
\newblock Fast approximate quadratic programming for graph matching.
\newblock \emph{PLOS one}, 10\penalty0 (4):\penalty0 e0121002, 2015.

\bibitem[Vuli{\'c} et~al.(2020)Vuli{\'c}, Ruder, and S{\o}gaard]{vulic2020all}
Ivan Vuli{\'c}, Sebastian Ruder, and Anders S{\o}gaard.
\newblock Are all good word vector spaces isomorphic?
\newblock \emph{arXiv preprint arXiv:2004.04070}, 2020.

\bibitem[Woo et~al.(2023)Woo, Debnath, Hu, Chen, Liu, Kweon, and Xie]{woo2023convnext}
Sanghyun Woo, Shoubhik Debnath, Ronghang Hu, Xinlei Chen, Zhuang Liu, In~So Kweon, and Saining Xie.
\newblock Convnext v2: Co-designing and scaling convnets with masked autoencoders.
\newblock In \emph{Proceedings of the IEEE/CVF Conference on Computer Vision and Pattern Recognition}, pages 16133--16142, 2023.

\bibitem[Wu et~al.(2020)Wu, Belinkov, Sajjad, Durrani, Dalvi, and Glass]{wu2020similarity}
John~M Wu, Yonatan Belinkov, Hassan Sajjad, Nadir Durrani, Fahim Dalvi, and James Glass.
\newblock Similarity analysis of contextual word representation models.
\newblock \emph{arXiv preprint arXiv:2005.01172}, 2020.

\end{thebibliography}
}


% \clearpage
% \setcounter{page}{1}
\maketitlesupplementary

\appendix

\setcounter{table}{0}
\setcounter{figure}{0}
\renewcommand{\thetable}{A.\arabic{table}}
\renewcommand{\thefigure}{A.\arabic{figure}}

\section{Varying the Number of Samples}
\label{A_sec:ablation_base_query}
\begin{figure}
    \centering
    \includegraphics[width=1\linewidth]{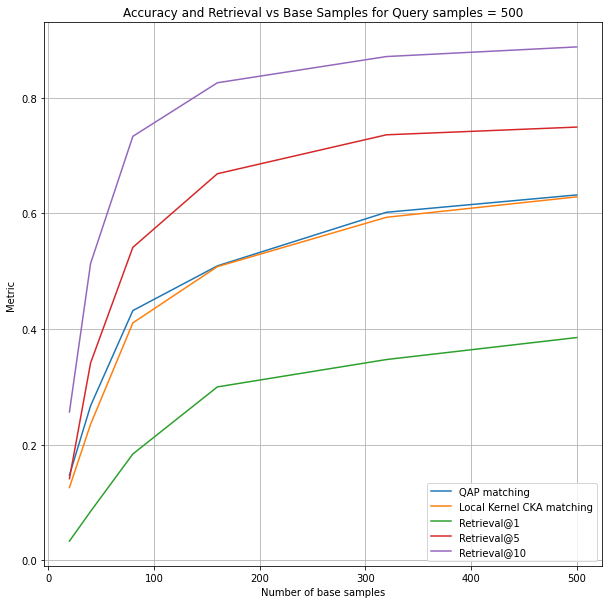}
    \caption{\textbf{Accuracy and Retrieval Scores} of QAP Matching and Local CKA-based retrieval as the number of base samples is varied, keeping the number of query samples fixed at 500.}
    \label{fig:acc_vs_base}
\end{figure}

\begin{figure}
    \centering
    \includegraphics[width=1\linewidth]{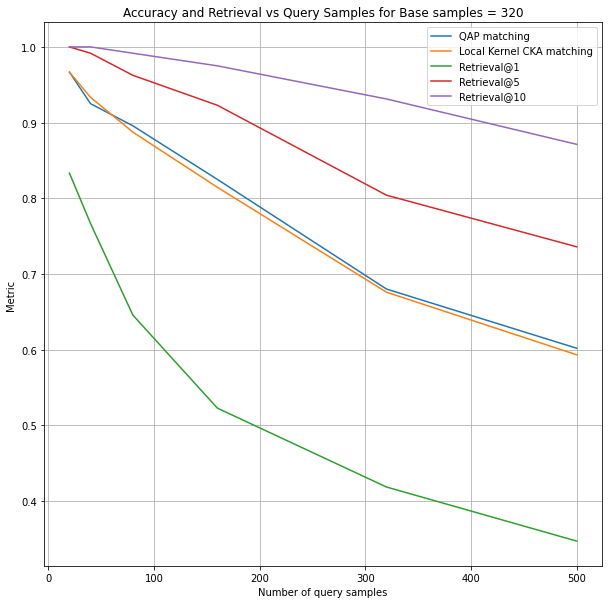}
    \caption{\textbf{Accuracy and Retrieval Scores} of QAP Matching and Local CKA based retrieval as the number of query samples is varied, keeping the number of base samples fixed at 320.}
    \label{fig:acc_vs_query}
\end{figure}

In Figure \ref{fig:acc_vs_base}, we show QAP and local CKA matching accuracies and retrieval scores for different number of base samples $M$, keeping the number of query samples $N$ constant at 500. It can be observed that as $M$ increases, accuracy/retrieval scores improve, demonstrating the importance of seed initialization for matching algorithms. Figure \ref{fig:acc_vs_query} shows the accuracy/retrieval scores as $N$ the number of query samples changes keeping the number of base samples constant at M=320. We see that QAP matching accuracy as local CKA-based retrieval scores decrease with increase in $N$, but we still get 70\% matching accuracy when $\frac{M}{N}=1$.

\begin{table*}[t]
    \centering
    \caption{\textbf{Image Encoders Summary.} List of hugging face vision encoder names and information regarding their train data, paradigm, dataset size, model type, and model sizes for the comparison in Figure \ref{fig:image_vs_text_encoders_cka}  and Table \ref{tab:vision_text_cka_combos}.\vspace{-0.2cm}}
    
    \adjustbox{width=\textwidth}{%
  \begin{tabular}{lccccc}
    \toprule
    \textbf{Model Name} & \textbf{Training Data} & \textbf{Training Paradigm} & \textbf{Model Type} & \textbf{Training Data Size} & \textbf{Model Size} \\
    \midrule
    facebook\textbackslash dino-vits8 & ImageNet-1k & DinoV1 & vit-small & 1.2 & 22 \\
    openai\textbackslash clip-vit-large-patch14-336 & CLIP-400M & Language Supervised & vit-large & 400 & 307 \\
    facebook\textbackslash dinov2-base & LVD-142M & DinoV2 & vit-base & 142 & 86 \\
    facebook\textbackslash dinov2-small & LVD-142M & DinoV2 & vit-small & 142 & 22 \\
    facebook\textbackslash dinov2-large & LVD-142M & DinoV2 & vit-large & 142 & 307 \\
    facebook\textbackslash dinov2-giant & LVD-142M & DinoV2 & vit-giant & 142 & 1000 \\
    openai\textbackslash clip-vit-base-patch16 & CLIP-400M & Language Supervised & vit-base & 400 & 86 \\
    facebook\textbackslash dino-vitb8 & ImageNet-1k & DinoV1 & vit-base & 1.2 & 86 \\
    timm\textbackslash convnext\_base.fb\_in1k & ImageNet-1k & Supervised & convnext-base & 1.2 & 89 \\
    timm\textbackslash convnext\_tiny.fb\_in1k & ImageNet-1k & Supervised & convnext-tiny & 1.2 & 29 \\
    facebook\textbackslash convnext-base-224-22k & ImageNet-21k & Supervised & convnext-base & 14.1 & 89 \\
    timm\textbackslash convnext\_base.fb\_in22k & ImageNet-21k & Supervised & convnext-base & 14.1 & 89 \\
    timm\textbackslash vit\_base\_patch16\_224.augreg\_in21k & ImageNet-21k & Supervised & vit-base & 14.1 & 86 \\
    timm\textbackslash vit\_small\_patch16\_224.augreg\_in1k & ImageNet-1k & Supervised & vit-small & 1.2 & 22 \\
    \bottomrule
    \end{tabular}
    }
    \label{tab:vision_list}
\end{table*}

\begin{table*}[t]
    \centering
       \caption{\textbf{Text Encoders Summary.} List of huggingface text encoder names and information regarding their train data, paradigm, dataset size, and model sizes for the comparison in Figure \ref{fig:image_vs_text_encoders_cka}  and Table \ref{tab:vision_text_cka_combos} \vspace{-0.2cm}}
    \label{tab:text_list}

    \adjustbox{width=\textwidth}{%
\begin{tabular}{lcccc}
\toprule
\textbf{Model Name} & \textbf{Model Size} & \textbf{Train Data} & \textbf{Training Paradigm} & \textbf{Training Data Size} \\
\midrule
all-mpnet-base-v1 & 109 & multiple datasets & contr. sent. & 1.12B sent. pairs \\
gtr-t5-base & 110 & multiple datasets & contr. sent. & 2B sent. pairs \\
paraphrase-MiniLM-L12-v2 & 33 & multiple datasets & contr. sent. & 10M sent. pairs \\
gtr-t5-large & 335 & multiple datasets & contr. sent. & 2B sent. pairs \\
all-mpnet-base-v2 & 109 & multiple datasets & contr. sent. & 1.12B sent. pairs \\
average\_word\_embeddings\_komninos & 66 & Wiki2015 & skipgram & 2 billion words \\
average\_word\_embeddings\_glove.6B.300d & 120 & Wiki2014, GigaWord 5 & glove & 6 billion tokens \\
all-MiniLM-L12-v1 & 33 & multiple datasets & contr. sent. & 1B sent. pairs \\
openai\_clip-vit-large-patch14 & 123 & CLIP-400M & contr. img-text & 400M image-text pairs \\
all-MiniLM-L12-v2 & 33 & multiple datasets & contr. sent. & 1B sent. pairs \\
all-MiniLM-L6-v2 & 22 & multiple datasets & contr. sent. & 1B sent. pairs \\
sentence-t5-base & 110 & multiple datasets & contr. sent. & 2B sent. pairs \\
msmarco-distilbert-dot-v5 & 66 & MSMarco & contr. sent. & 500k sent. pairs \\
paraphrase-MiniLM-L3-v2 & 17 & multiple datasets & contr. sent. & 10M sent. pairs \\
paraphrase-albert-small-v2 & 11 & multiple datasets & contr. sent. & 10M sent. pairs \\
all-MiniLM-L6-v1 & 22 & multiple datasets & contr. sent. & 1B sent. pairs \\
all-distilroberta-v1 & 82 & OpenWebTextCorpus & contr. sent. & 1B sent. pairs \\
sentence-t5-large & 335 & multiple datasets & contr. sent. & 2B sent. pairs \\
All-Roberta-large-v1 & 355 & multiple datasets & contr. sent. & 1B sent. pairs \\
msmarco-bert-base-dot-v5 & 109 & MSMarco & contr. sent. & 500k sent. pairs \\
sentence-t5-xxl & 4870 & multiple datasets & contr. sent. & 2B sent. pairs \\
paraphrase-TinyBERT-L6-v2 & 66 & multiple datasets & contr. sent. & 10M sent. pairs \\
sentence-t5-xl & 1240 & multiple datasets & contr. sent. & 2B sent. pairs \\
gtr-t5-xxl & 4870 & multiple datasets & contr. sent. & 2B sent. pairs \\
paraphrase-distilroberta-base-v2 & 82 & multiple datasets & contr. sent. & 10M sent. pairs \\
gtr-t5-xl & 1240 & multiple datasets & contr. sent. & 2B sent. pairs \\
\bottomrule

\end{tabular}
}
\end{table*}

% We report the kernel CKA of several important CKA combinations in Table. \ref{tab:vision_text_cka_combos}. We notice that the best Text encoder that has been trained using only text information is allrobertalargev1 with dinoV2 with a CKA of 0.706. Hence we use allrobertalargev1 as the text encoder for all tasks and experiments in the main paper except for the cross-lingual experiments. For cross-lingual experiments we find that the best-performing text encoder is paraphrase-multilingual-mpnet-base-v2.  In Figure \ref{fig:image_vs_text_encoders_cka} we plot the CKA vs text model size for different vision encoder types, training paradigms and sizes. We see that higher CKA. We find that text model size is not very important for high CKA with the vision model and that more well-trained vision models on big datasets have high kernel CKA with text encoders irrespective of text model size. For example, language-supervised models(green) and DinoV2 models are all trained on datasets of the order the hundreds of millions of instances (LVD\-142's 142 million images and CLIP\-400M's 400 million image, caption pairs) and exhibit high CKA with language encoders of all sizes.

\section{Vision and Text Encoders}
\label{A_sec:vision_text}
% We measure CKA on combinations of a wide variety of vision and text encoders to study the effects of different vision and text encoder model sizes, training dataset sizes, and training paradigms on vision-language aligned as well as determine the optimal unaligned vision and text encoder to use for our caption-matching tasks. We make use of huggingface's transformers library for the vision models while  sentence transformers library for the text encoders. In table \ref{tab:vision_list} we go over the vision models, their train data, paradigms as well as model types and sizes, and in table \ref{tab:text_list} we do the same for the different ext encoders.For vision models, we cover 4 training paradigms -supervised, dinov1, dinov2 and language supervised as well as training dataset size ranging from 1 million to 140 million. For text encoders we mainly use sentence transformers that have been trained for semantic search using a constastive sentence pairs loss. The dataset sizes range from 500k to 2B. 

CKA is measured on combinations of a wide variety of vision and text encoders to examine the impact of: model sizes, dataset regimes, and training paradigms on vision-language alignment. This analysis also identifies the optimal pair of unaligned vision and text encoder for caption-matching tasks. Huggingface's transformers library is utilized for vision models, while the sentence transformers library is employed for text encoders. Table \ref{tab:vision_list} details the vision models, their training data, paradigms, and model types and sizes. Similarly, Table \ref{tab:text_list} presents information on various text encoders. The study covers three training paradigms for vision models: supervised, self-supervised, and language-supervised, with training dataset sizes ranging from 1 million to 400 million images. Text encoders predominantly use sentence transformers, trained for semantic search using a contrastive sentence pairs loss, with dataset sizes varying from 500k to 2B.

Kernel CKA of various model combinations is presented in Table \ref{tab:vision_text_cka_combos}. The top-performing text encoder trained exclusively on text information is identified as All-Roberta-large-v1 paired with DINOv2, achieving a CKA of 0.706. Consequently, All-Roberta-large-v1 is selected as the text encoder for all tasks and experiments in the main paper, except for cross-lingual experiments. For these, paraphrase-multilingual-mpnet-base-v2 emerges as the most effective text encoder.

\begin{figure}[t]
    \centering
    \includegraphics[width=1\linewidth]{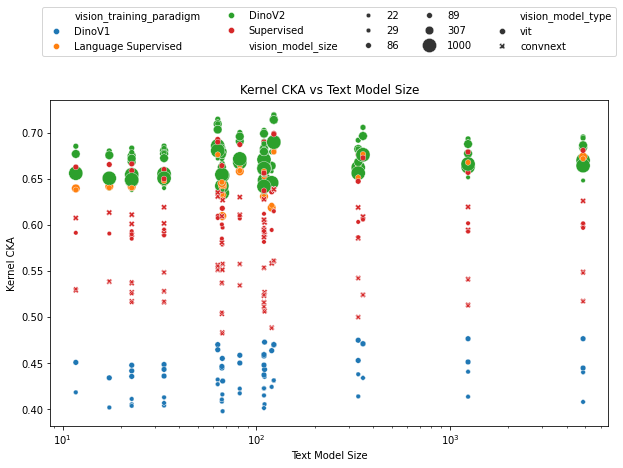}
    \caption{\textbf{CKA \textit{vs.} text model size} for vision encoders of different training paradigms, model types, and model sizes. We see that text model size is not the most important for high semantic similarity with vision models.}
    \label{fig:image_vs_text_encoders_cka}
\end{figure}

Figure \ref{fig:image_vs_text_encoders_cka} illustrates the relationship between CKA and text model size across different vision encoder types, training paradigms, and sizes. It is observed that text model size has a limited impact on achieving high CKA with the vision model. Well-trained vision models on large datasets consistently show high kernel CKA with text encoders, regardless of text model size. For instance, language-supervised models (green) and DINOv2 models, which are trained on datasets with hundreds of millions of instances (such as LVD-142’s 142 million images and CLIP-400M’s 400 million image-caption pairs), demonstrate high CKA with language encoders of various sizes.

\section{Layerwise CKA Analysis}
\label{A_sec:layerwise_cka}
\begin{figure}[t]
    \begin{subfigure}{.45\columnwidth}
        \includegraphics[width=\linewidth]{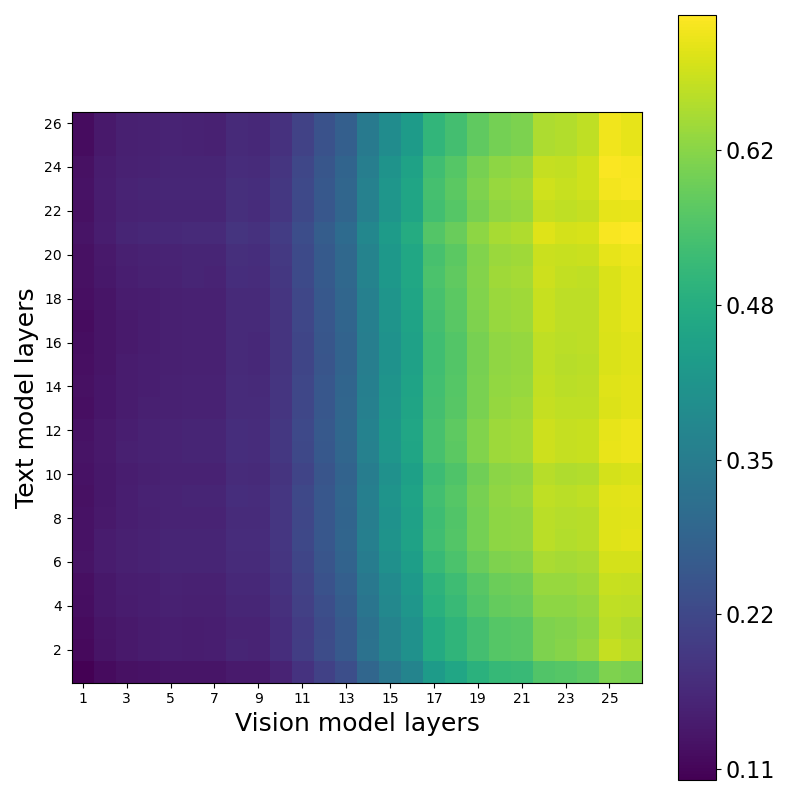}
        \caption{}
    \end{subfigure}%
    \begin{subfigure}{.45\columnwidth}
        \includegraphics[width=\linewidth]{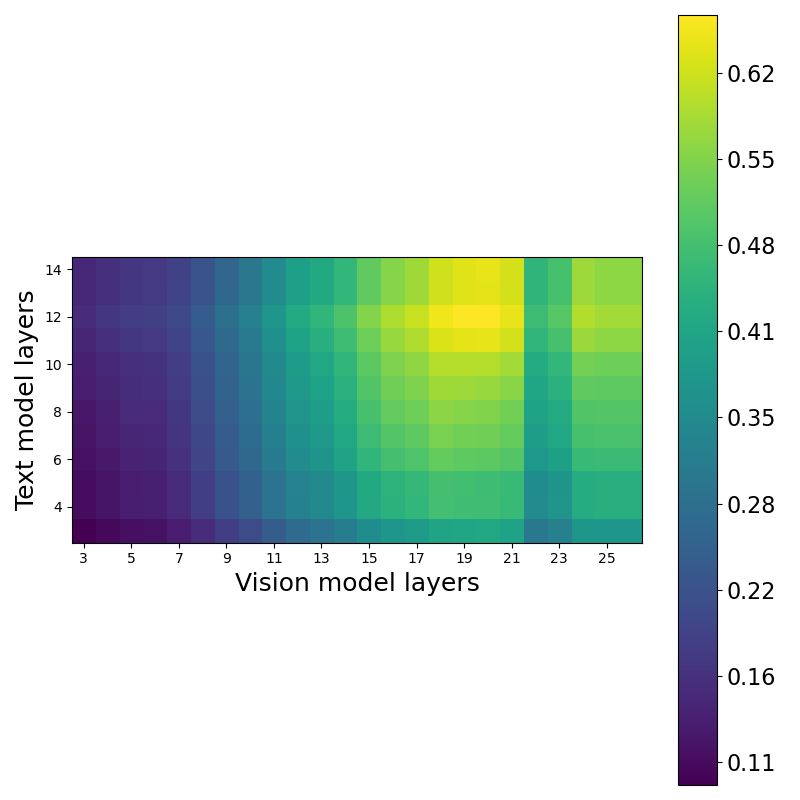}
        \caption{}
    \end{subfigure}%
    \caption{\textbf{Layer-wise CKA heatmap illustration.} The heatmaps depict the CKA scores obtained by varying the layers from which the text and visual embeddings are taken. \textbf{On the left:} CKA scores for  All-Roberta-large-v1 and DINOv2 unaligned combination. \textbf{On the right:} CKA scores for CLIP text and vision encoders. In both cases, we observe that the CKA scores are low for earlier layer embeddings of the vision model and they improve when the embeddings later layers are considered. This illustrates that both aligned and unaligned text-vision encoders behave similarly in terms of the cross-modal similarity \wrt CKA.}
    \label{fig:layerwise}
\end{figure}

Figure \ref{fig:layerwise}, Table \ref{tab:qap_clip}, and Table \ref{tab:qap_dino_bert} show the progression of CKA and QAP matching scores across layers for both text and vision models. We explore two configurations: one involves comparing layers of All-Roberta-large-V1 and DINOv2 VIT-L/14, while the other examines layers of CLIP's vision and text hidden states. For CLIP, the layer $proj$ points to the final image and text embeddings that were passed through the final projection layers. In the first configuration, CKA and QAP scores gradually improve where the image model layer has a far greater effect on the similarity than the text model layer. On the other hand, the second configuration reveals that the QAP matching score in CLIP manifests prominently in the absolute last layers of both the vision/text encoders.

As shown in Table \ref{tab:qap_clip}, the CLIP model obtains a significant jump in matching score after the projection head, highlighting the central role of this layer in aligning text and image modalities within a unified representation space. Here, the QAP matching accuracy does not follow a linear increase over the layers for CLIP, but rather suddenly jumps from 0.29 to 0.79 from the last layer to the projection head. This likely suggests that most of the CLIP performance comes from the projection heads ensuring a high statistical similarity. 
In contrast, Table \ref{tab:qap_dino_bert} shows that DINOv2 and All-Roberta-large-v1 demonstrate a consistent improvement in the matching accuracy across successive layers, suggesting an inherent alignment process within their architectures in a hierarchical way. Here, the QAP matching accuracy linearly increases for the DINOv2 and All-Roberta-large-v1 combination when we fix the last layer of All-Roberta-large-v1 and vary the layers of DINOv2. Inversely, when we fix the last layer of DINOv2 and vary the layers of the text encoder, the QAP starts high at 0.44 and reaches 0.68 at the top layer, thus, we hypothesize that the text encoder representations do not change as much as the image representations.

% Furthermore, from Table \ref{tab:qap_clip}, Differently, from Table \ref{tab:qap_dino_bert}, 

% \begin{figure}
%     \centering
%     \includegraphics[width=\columnwidth]{images/clip_vs_dino-cka_layerwise.png}
%     \caption{Layer-wise CKA heatmap}
%     \label{fig:layerwise}
% \end{figure}

\begin{table}[t]
    \centering
        \caption{\textbf{QAP accuracy for different layers} of vision and text encoder of CLIP model.\vspace{-0.2cm}}
    \label{tab:qap_clip}
    \adjustbox{width=1\columnwidth}{
    \begin{tabular}{l|r|rrrrrr}
    \toprule
        & & \multicolumn{5}{c}{Vision} \\
        \cmidrule(lr){3-8}
        & & 6th & 11th & 16th & 21st & 26th & proj\\
        \midrule
        \multirow{5}{*}{Text} 
        & 6th &  0.02 &  0.022 &  0.022 &  0.098 &  0.126 & 0.118\\
        & 11th &  0.028 &  0.038 &  0.016 &  0.248 &  0.278 & 0.278 \\
        & 14th &  0.026 &  0.03 &  0.036 &  0.238 &  0.282 & 0.296 \\
        & proj & 0.038 &  0.026 &  0.034 &  0.622 &  0.716 & 0.792 \\
        \bottomrule
    \end{tabular}
    }
\end{table}

\begin{table}[t]
    \centering
    \caption{\textbf{QAP accuracy for different layers} of DINOv2 and All-Roberta-large-v1 models.\vspace{-0.2cm}}
    \label{tab:qap_dino_bert}
    \begin{tabular}{l|r|rrrrr}
        \toprule
        & & \multicolumn{5}{c}{Vision} \\
        \cmidrule(lr){3-7}
        & & 6th & 11th & 16th & 21st & 26th \\
        \midrule
        \multirow{5}{*}{Text} 
        & 6th &  0.008 &  0.020 &  0.150 &  0.314 &  0.448 \\
        & 11th &  0.010 &  0.022 &  0.146 &  0.360 &  0.498 \\
        & 16th &  0.008 &  0.016 &  0.194 &  0.334 &  0.500 \\
        & 21st &  0.002 &  0.004 &  0.148 &  0.420 &  0.538 \\
        & 26th &  0.008 &  0.016 &  0.198 &  0.450 &  0.672 \\
        \bottomrule
    \end{tabular}
\end{table}

\begin{table*}[t]
    \centering
    \caption{\textbf{Impact of adding noise to the embeddings}. Performance comparison, in terms of matching accuracy, between relative representations~\cite{moschella2022relative} and our global CKA-based QAP approach is shown for the image-caption matching task with 320 base samples and 500 query samples on COCO validation set. Gaussian noise with std-dev ($\sigma$) being a multiple of the embeddings std-dev is added to both image and textual embeddings. Noise level of 0 ($\sigma=0$) denotes the performance for the original embeddings. The relative performance drop for a noise level from its reference ($\sigma=0$) is shown in parenthesis. In comparison to relative representations, our QAP approach performance drops at a slower rate as $\sigma$ increases, illustrating better noise robustness for our approach.\vspace{-0.2cm}}
    \label{tab:noise_ablation}
    
    \adjustbox{width=1.7\columnwidth}{
    \begin{tabular}{l|cccccc}
    \toprule
         \multirow{2}{*}{\textbf{Method}} & \multicolumn{6}{c}{\textbf{Noise Level ($\sigma$)}}\\
         & 0.0 & 0.1 & 0.2 & 0.3 & 0.4 & 0.5   \\
        \midrule         
        Relative representations~\cite{moschella2022relative} & 47.3 & 45.3 ({\small{\textcolor{teal}{$\mathord{\downarrow}4.4$}}}) & 44.2 ({\small{\textcolor{teal}{$\mathord{\downarrow}6.5$}}}) & 41.3 ({\small{\textcolor{teal}{$\mathord{\downarrow}12.7$}}}) & 39.0 ({\small{\textcolor{teal}{$\mathord{\downarrow}17.6$}}}) & 35.6 ({\small{\textcolor{teal}{$\mathord{\downarrow}24.8$}}}) \\
        \textbf{Ours (QAP)} & 53.9 & 53.7 ({\small{\textcolor{teal}{$\mathord{\downarrow}0.3$}}}) & 51.8 ({\small{\textcolor{teal}{$\mathord{\downarrow}3.9$}}}) & 48.7 ({\small{\textcolor{teal}{$\mathord{\downarrow}9.5$}}}) & 46.9 ({\small{\textcolor{teal}{$\mathord{\downarrow}13.0$}}}) & 43.3 ({\small{\textcolor{teal}{$\mathord{\downarrow}19.6$}}}) \\
        \bottomrule
    \end{tabular}
    } 
\end{table*}

\section{Mathematical Relationship between Local CKA-based Retrieval and Relative Representations}
\label{A_sec:math_relative}

In this section, we provide derivations that show that the relative representations method \cite{moschella2022relative} can be seen as a particular case of our proposed $\localcka$ method.
Denote the set of query and base representations samples respectively as $\rmQ_A = \left[ \vq_1^A, \ldots, \vq_N^A \right] \in \sR^{d_A \times N}$ and $\rmB_A = \left[ \vb_1^A, \ldots, \vb_M^A \right] \in \sR^{d_A \times M}$, where  $A \in \{ I, C \}$ for images and captions, the retrieval matrix for the relative representations (RR) method is therefore given by:
\begin{align*}
    \rmR^{\text{RR}} = \rmQ_I^\top \rmB_I \rmB_C^\top \rmQ_C \in \sR^{N\times N}.
\end{align*}
From which, for instance, the $i$-th image query is mapped to its corresponding caption via:
\begin{align}
    \argmax_j \emR_{ij}^{\text{RR}} = \argmax_j \, (\vq_i^I)^\top \rmB_I \rmB_C^\top \vq_j^C.
\end{align}
Whereas, our proposed $\localcka$ method constructs the retrieval matrix $\rmR^{\text{Ours}}$ having entries $\emR_{ij}^{\text{Ours}} = \localcka\left( \vq_i^I, \vq_j^C \right)$ with:
\begin{align}
     \localcka\left( \vq_i^I, \vq_j^C \right) = \cka \left( \rmK_{[\rmB_I, \vq_i^I]}, \rmK_{[\rmB_C, \vq_j^C]} \right).
\end{align}
In particular, taking the particular case of the linear kernel and defining the CKA score as the trace of the product of two kernels, i.e., $\cka(\rmK, \rmL) = \tr \left( \rmK \rmL \right)$. We first have, for $A\in \{I, C \}$:
\begin{align*}
    \rmK_{[\rmB_A, \vq_i^A]} =  [\rmB_A, \vq_i^A]^\top [\rmB_A, \vq_i^A] = \begin{bmatrix}
        \rmB_A^\top \rmB_A & \rmB_A^\top \vq_i^A \\
        \left( \rmB_A^\top \vq_i^A \right)^\top & \Vert \vq_i^A \Vert^2
    \end{bmatrix}.
\end{align*}
Hence, we have:
\begin{align*}
    \tr &\left( \rmK_{[\rmB_I, \vq_i^I]} \rmK_{[\rmB_C, \vq_j^C]} \right) = \tr \left( \rmB_I^\top \rmB_I \rmB_C^\top \rmB_C \right) \\
    &+ 2 \underbrace{\left( \vq_i^I \right)^\top \rmB_I \rmB_C^\top \vq_j^C}_{\text{relative representations term}} + \Vert \vq_i^I \Vert^2 \Vert \vq_j^C \Vert^2.
\end{align*}
Therefore, in this particular case, there is equivalence between our method and the relative representations method, since $\emR_{ij}^{\text{Ours}} = \emR_{ij}^{\text{RR}} + c$ where $c$ is a constant scalar if the representations are normalized. As such, the relative representations method falls within our proposed $\localcka$ method if one considers the linear kernel and takes the trace instead of the $\hsic$ metric. Therefore, our proposed method is more general since it relies on general kernel functions and the $\hsic$ metric, which might explain its performance.

\noindent\textbf{Impact of noise addition:} Table~\ref{tab:noise_ablation} shows the performance comparison between relative representations~\cite{moschella2022relative} and our global CKA-based QAP approach for the image-caption matching task with 320 base samples and 500 query samples on COCO validation set. For this experiment, 10 trials were conducted with different seeds and clustering of base samples was employed. Gaussian noise with std-dev ($\sigma$) being a multiple of the embeddings std-dev is added to both image and textual embeddings. The performance of original embeddings is also shown for reference (noise level of 0, \ie, $\sigma=0$). The relative performance drop for a noise level from its reference ($\sigma=0$) is shown in parenthesis. Compared to relative representations, our QAP approach performance drops at a slower rate as $\sigma$ increases. \Eg, for $\sigma=0.2$, relative representations matching accuracy drops 6.5\% from it maximum of 47.3, while ours is more robust and drops only 3.9\% from its maximum of 53.9 when $\sigma=0$. These results show that our QAP approach is more robust to noise addition, in comparison to relative representations.

\section{Other text encoders}
\label{A_sec:text_encoders}
Evaluating on COCO with M=320 and N=500, Table \ref{tab:text_encoders} shows that DINOv2-large achieves high QAP accuracy and retrieval performance when combined with different text encoders.
% \eg, paraphrase-mpnet-base-v2 exhibits superior retrieval performance. 
This underscores the potential of pairing well-trained sentence and vision encoders for achieving high semantic similarity between image and text embeddings

\begin{table}[h]
    \centering
    \caption{Comparison of CKA, QAP acc. and local CKA retrieval for different text encoders with DINOv2-large image encoder.\vspace{-0.25cm}}
    \adjustbox{width=0.85\columnwidth}{
\begin{tabular}{lccc}
\hline
                            Text Encoder &  Kernel CKA &  QAP Acc. &  Ret @ 5 \\
\hline
            all-roberta-large-v1 &       0.690 &    64.93 & 77.27 \\
paraphrase-distilroberta-base-v2 &       0.689 &    65.07 & 76.33 \\
        paraphrase-mpnet-base-v2 &       0.695 &    68.20 & 81.07 \\
               sentence-t5-large &       0.660 &    57.87 & 69.13 \\
                 sentence-t5-xxl &       0.677 &    63.40 & 73.00 \\
\hline
\end{tabular}
}
\vspace{-0.3cm}
    \label{tab:text_encoders}
\end{table}

\section{Simple projection} 
We trained a 2-layer MLP on frozen DINOv2-large encoder till convergence using CLIP loss and MSE loss. For fair comparison with our setting, we use 320 training and 500 query image-text samples. Results in Table \ref{tab:proj} are averaged over 3 seeds. Notably, QAP matching and local-CKA retrieval excel over projection learning, which demands hyperparameter tuning. In contrast, QAP and local-CKA provide a novel, training-free mechanism to evaluate encoder representational similarity, demonstrating effective latent space communication.

\section{Effect of unimodal tasks on alignment}
\label{A_sec:unimodal_tasks}
Table \ref{tab:tasks} shows using ViT, DETR, DPT, and SegFormer vision encoders for local-CKA and QAP matching on COCO captions (M=320, N=500). ViT is trained on ImageNet-1k (classification), DETR on COCO 2017 (detection), DPT on 1.4M depth images (depth estimation), and SegFormer is fine-tuned on ADE20k (semantic segmentation). Results indicate that classification models exhibit higher semantic similarity to all-roberta-large text encoder in QAP accuracy and local-CKA scores than pixel-level tasks such as object detection, segmentation, and depth estimation.

\begin{table}[h]
\vspace{-0.1cm}
    \centering
    \begin{minipage}[t]{0.45\columnwidth}
        \centering
        \caption{QAP acc. and Top-5 retrieval scores  on COCO.\vspace{-0.2cm}}
        \resizebox{\columnwidth}{!}{
        \begin{tabular}{lcc}
            \hline
            Method & QAP acc & Ret @ 5 \\
            \hline
            Proj. + MSE & 59.8 & 73.0 \\
            Proj. + CLIP & 55.4 & 68.1 \\
            QAP & \textbf{65.9} & - \\
            Local CKA & 64.3 & \textbf{76.0} \\
            \hline
        \end{tabular}
        }
        
        \label{tab:proj}
    \end{minipage}
    \hfill % Optional; for spacing
    \begin{minipage}[t]{0.47\columnwidth}
        \centering
        \caption{Unimodal tasks' effect on image-text alignment.\vspace{-0.2cm}}
        \resizebox{\columnwidth}{!}{
        \begin{tabular}{l|cc}
            \hline
            Vision model & QAP acc & Ret @ 5 \\
            \hline
            ViT & 35.3 & 56.1 \\
            DETR & 26.5 & 39.8 \\
            DPT & 22.7 & 34.1 \\
            Segformer & 16.8 & 33.4 \\
            \hline
        \end{tabular}
        }
        
        \label{tab:tasks}
    \end{minipage}
    \vspace{-0.1cm}
\end{table}

\section{Additional Retrieval Results}
\label{A_sec:caps_ret}
While the performance on the image retrieval task was reported in Table 2 of the main manuscript, here in Table \ref{tab:caption_reverse}, we show the NoCaps and Coco caption retrieval results in the reverse setting. In this configuration, the retrieving objective shifts to finding the correct caption from a pool of $N$ captions when given a single image. The matching objective remains consistent, but, instead of shuffling the captions, the images themselves are shuffled. While the matching accuracies express minimal changes in this setting, the retrieval accuracies display notable discrepancies.

A plausible explanation for the reduced retrieval scores associated with the relative representation method is the heightened semantic variability inherent in the image domain compared to the caption domain. A considerable number of images share very similar captions, leading to a compressed semantic space for the captions. Consequently, caption embeddings become more closer to one another, making the retrieval a lot harder.

\begin{table*}[]
    \centering
    \caption{\textbf{Reverse Caption Retrieval Results for COCO and NoCaps}. In this setting, the retrieval objective is, given one image, to retrieve the correct caption from the overall set of $N$ captions. The matching objective remains quite similar but instead of shuffling the captions, this time, the images are shuffled.\vspace{-0.2cm}}
    \begin{tabular}{ll|cc|cc}
    \toprule[0.1em]
    \multirow{2}{*}{\textbf{Method}}  &   \multirow{2}{*}{\textbf{ Vision Model}} & \multicolumn{2}{c|}{\textbf{NoCaps}~\cite{agrawal2019nocaps}} & \multicolumn{2}{c}{\textbf{COCO}~\cite{mscoco}} \\
    & & Matching accuracy &  Top-5 retrieval &  Matching accuracy &  Top-5 retrieval  \\
    \specialrule{1pt}{1pt}{1pt}%
    Cosine Similarity* & CLIP~\cite{radford2021learning} &  \demph{99.5} &  \demph{99.6} & \demph{97.1} & \demph{98.5} \\
    \specialrule{0.5pt}{0.5pt}{0.5pt}%
    \multirow{3}{*}{Linear regression} & CLIP-V~\cite{radford2021learning} & 63.6 & 70.1 & 72.6 & 83.9 \\
             &  ConvNeXt~\cite{woo2023convnext} & 22.8 & 38.9 & 43.8 & 65.7 \\
             &    DINOv2~\cite{oquab2023dinov2} & 46.8 & 59.9 & 56.2 & 75.9 \\
    \specialrule{0.5pt}{0.5pt}{0.5pt}%
    \multirow{2}{*}{Relative}  & CLIP-V~\cite{radford2021learning} & 61.3 & 3.0 & 61.6 & 2.9 \\
      \multirow{2}{*}{representations~\cite{moschella2022relative}} &  ConvNeXt~\cite{woo2023convnext} &  25.5 & 2.7 & 38.6 & 12.9 \\
             &    DINOv2~\cite{oquab2023dinov2} &  45.9 & 38.1 & 47.7 & 43.7 \\
    \specialrule{0.5pt}{0.5pt}{0.5pt}%
    \multirow{3}{*}{\textbf{Ours: QAP}} & CLIP-V~\cite{radford2021learning} & 67.3 & - & 72.8 & - \\
             &  ConvNeXt~\cite{woo2023convnext} &  45.9 & - & 65.1 & - \\
             &    DINOv2~\cite{oquab2023dinov2} &  58.5 & - & 65.9 & - \\
    \specialrule{0.5pt}{0.5pt}{0.5pt}%
    \multirow{3}{*}{\textbf{Ours: Local CKA}} & CLIP-V~\cite{radford2021learning} & 65.1 & 65.9 & 71.9 & 80.5 \\
             &  ConvNeXt~\cite{woo2023convnext} &  44.8 & 33.0 & 63.8 & 74.3 \\
             &    DINOv2~\cite{oquab2023dinov2} &  55.7 & 64.2 & 64.3 & 76.0\\
    \bottomrule[0.1em]
    \end{tabular}
    
    \label{tab:caption_reverse}
\end{table*}

\section{Additional Cross-Lingual Matching Results}
\label{A_sec:cross_caps_ret}

For completeness, we report the results in Table \ref{tab:cross-lingual-reverse} for the reverse setting of the cross-lingual image caption matching/retrieval task mentioned in the main paper. Given $N$ captions in say, German, and $N$ shuffled images the objective is to match each German caption with the correct image. In retrieval, the goal is to select the most fitting image from the retrieval set given a German caption. We notice that the matching accuracies remain the same as the direction doesn't affect the matching. However, in the case of reverse retrieval, we notice that CLIP's retrieval@5 drops by over 4.5\% on average when compared to our local CKA based retrieval of 2.1\%. 

In Table \ref{tab:cross_lingual_specific} we report the results for when we use language-specific BERT Sentence encoders for the cross-lingual caption matching/ retrieval task for 5 languages. For all these cases, the vision encoder is kept fixed as OpenAI's CLIP-VIT-L-14 trained on English image, caption pairs. We notice that the semantic alignment with the vision encoder in terms of CKA as well as matching/retrieval performance drops with language-specific encoders when compared to using a multi-lingual model like multilingual-mpnet-base-v2. We believe this could be due to the multi-lingual model being trained on a lot more data in comparison to the language-specific ones thus resulting in more meaningful embedding spaces.

\begin{table*}[h]
    \centering
    \caption{\textbf{Cross-Lingual image matching and retrieval performance comparison. Here we use multilingual captions to retrieve images from the COCO validation set.} Using QAP and local CKA-based methods we are able to do cross-lingual image matching/retrieval using CLIP's ViT-L vision encoder and a multi-lingual sentence transformer paraphrase-multilingual-mpnet-base-v2. While CLIP performs well on the Latin languages, it degrades on non-Latin languages. In comparison, our QAP and Local-CKA-based methods perform comparably in Latin languages while outperforming non-Latin languages, highlighting the efficacy of our training-free transfer approach.\vspace{-0.2cm}}
\begin{tabular}{llcc|cccc|ccc}
\toprule
 \multirow{2}{*}{\textbf{Language}} & & \multicolumn{2}{c|}{\textbf{Kernel CKA}}   &  \multicolumn{4}{c|}{\textbf{Matching Accuracy}} & \multicolumn{2}{c}{\textbf{Retrieval @ 5}}  \\
  &   &    CLIP               &     Ours        &      CLIP &  Relative\cite{moschella2022relative} &  Linear &  Ours (QAP)  &      CLIP  &  Ours (Local)           \\
\midrule

\multirow{5}{*}{\textbf{Latin}} 
     & de &            0.472 &       0.627 &      43.5 &          35.0 &        19.3 &     39.7 &              54.9 &  57.2 \\
     & en &            0.567 &       0.646 &      80.9 &          52.5 &        25.6 &     51.3 &              90.4 &  66.7 \\
     & es &            0.471 &       0.634 &      50.4 &          37.8 &        19.7 &     40.9 &              63.9 &  57.9 \\
     & fr &            0.477 &       0.624 &      50.8 &          37.5 &        18.8 &     40.3 &              65.9 &  56.9 \\
     & it &            0.472 &       0.638 &      41.9 &          37.2 &        19.7 &     38.7 &              52.9 &  57.0 \\

\midrule
 \multirow{6}{*}{\textbf{Non-Latin}} 

     & jp &            0.337 &       0.598 &      12.9 &          28.3 &        15.2 &     30.2 &              17.8 &  48.6 \\
     & ko &            0.154 &       0.620 &       0.9 &          30.4 &        15.3 &     31.3 &               2.2 &  48.4 \\
     & pl &            0.261 &       0.642 &       8.1 &          36.6 &        21.0 &     40.0 &              15.7 &  55.9 \\
     & ru &            0.077 &       0.632 &       1.7 &          31.8 &        16.3 &     34.8 &               3.5 &  53.9 \\
     & tr &            0.301 &       0.624 &       7.8 &          35.8 &        18.7 &     38.9 &              14.6 &  53.1 \\
     & zh &            0.133 &       0.641 &       2.4 &          36.5 &        19.2 &     39.9 &               4.8 &  53.7 \\

   \midrule

   & \mycc \textbf{Avg.}       &     \mycc    --     & \mycc  --     &  \mycc 27.4    & \mycc  36.3      &  \mycc 18.9  & \mycc \textbf{38.7} &  \mycc 35.1 &  \mycc \textbf{55.4} \\
\bottomrule
\end{tabular}

    \label{tab:cross-lingual-reverse}
\end{table*}

\begin{table*}[]
    \centering
        \caption{\textbf{Language-specific encoders for cross-lingual caption matching/retrieval for 5 languages}. Language-specific encoders have less semantic similarity with the vision encoder in terms of CKA as well as poorer matching/accuracy performances when compared to multi-lingual models like multilingual-mpnet-base-v2 which is reported in Table 4.\vspace{-0.2cm}}
    \label{tab:cross_lingual_specific}
\adjustbox{width=\textwidth}{%
\begin{tabular}{llccccc}
    \toprule
    \textbf{Language} & \textbf{Language model} & \textbf{CKA} & \textbf{Linear} & \textbf{Relative} & \textbf{QAP} & \textbf{Retrieval@5} \\
    \midrule
    es & hiiamsid\textbackslash sentence\_similarity\_spanish\_es & 0.568 & 15.9 & 25.1 & 28.6 & 50.0 \\
    fr & dangvantuan\textbackslash sentence-camembert-large & 0.569 & 22.5 & 31.5 & 35.0 & 53.1 \\
    it & nickprock\textbackslash sentence-bert-base-italian-uncased & 0.543 & 16.0 & 22.0 & 26.4 & 47.8 \\
    jp & colorfulscoop\textbackslash sbert-base-ja & 0.457 & 9.2 & 12.1 & 14.5 & 33.7 \\
    tr & emrecan\textbackslash bert-base-turkish-cased-mean-nli-stsb-tr & 0.564 & 23.1 & 34.7 & 38.3 & 54.3 \\
    \bottomrule
    \end{tabular}
    }
\end{table*}

% \section{CLIP Score}
% \label{A_sec:clip_score}
% % debug this; remove it if not working @mayug

% Table \ref{tab:clip_score}. CLIP score comparison

% \begin{table}[]
% \begin{tabular}{lll}
% \toprule
% Method    & Matching Accuracy & CLIP Score \\
% \midrule
% Random    & 1/5               & 14.8       \\
% Relative  & 46.1              & 27.15      \\
% QAP       & 65.2              & 28.5       \\
% CLIP      & 94.8              & 30.14      \\
% All match & 100               & 30.15     \\
% \bottomrule
% \end{tabular}
% \caption{Caption}
% \label{tab:clip_score}

% \end{table}

\newcommand{\img}[1]{\includegraphics[width=3cm, height=3cm]{#1}}

\begin{table*}[t]
\centering 
\begin{tabular}{m{3cm}|m{3cm}|m{3cm}m{3cm}m{3cm}}
\multicolumn{1}{c}{\textbf{Original Image}} & \multicolumn{1}{c}{\textbf{Caption}} & \multicolumn{3}{c}{\textbf{Top-3 Retrieved Images}} \\
\toprule
\img{""} &  Two desktop computers sitting on top of a desk. &  \img{} &  \img{} &  \img{}  \\ 
\midrule
\img{""} &  A mother and baby elephant walking in green grass in front of a bond. &  \img{} &  \img{} &  \img{}  \\ 
\midrule
\img{""} &  a man is riding a surfboard at the beach &  \img{} &  \img{} &  \img{} \\ 
\midrule
\img{""} &  The Big Ben clock tower towering over the city of London. &  \img{} &  \img{} &  \img{} \\ 
\midrule
\img{""} &  A computer mouse is beside a notebook computer. &  \img{} &  \img{} &  \img{}\\ 
\bottomrule
\end{tabular}
\caption{\textbf{Local Kernel CKA Retrieval Mispredictions.} In accordance with the experimental protocol detailed in the main paper, we selected 320 base samples and conducted local Kernel CKA retrieval using an additional 500 query samples. Presented above are five example prediction retrievals for instances where the original image failed to secure a position within the top-5 retrievals. We observe that although the original image was not in the retrieved top-5, the retrieved images (top-3 shown here) closely resemble the corresponding caption, thereby highlighting the efficacy of our approach. }
\label{tab:image_examples}
\end{table*}

% \begin{sidewaystable*}
\begin{table*}[h]
    \centering
    \caption{\textbf{CKA for combinations of different vision and text encoders.} V, V\_tr, V\_tr\_size, V\_mod\_size stand for Vision model name, Vision train set, Vision train set size, and Vision model size respectively. T\_mod\_size stands for text model size. OpenAI's CLIP text encoder shows highest CKA with facebook dinoV2base closely followed by All-Roberta-large-v1. We make use of All-Roberta-large-v1 as the language encoder for all donwstream tasks and analysis in main text because All-Roberta-large-v1 has been trained using only text data and can be considered a purely textual encoder.\vspace{-0.2cm}}
    \label{tab:vision_text_cka_combos}
    \adjustbox{width=\textwidth}{%
   \begin{tabular}{llclllcc}
    \toprule
    V & T & CKA & V\_tr & V\_tr\_p & V\_tr\_size & V\_mod\_size & T\_mod\_size \\
    \midrule
    facebook\_dinov2-base & openai\_clip-vit-large-patch14 & 0.719 & LVD-142M & DinoV2 & 142 & 86 & 123 \\
    facebook\_dinov2-base & All-Roberta-large-v1 & 0.706 & LVD-142M & DinoV2 & 142 & 86 & 355 \\
    timm\_vit\_base\_patch16\_224.augreg\_in21k & openai\_clip-vit-large-patch14 & 0.698 & ImageNet-21k & Supervised & 14.1 & 86 & 123 \\
    facebook\_dinov2-large & sentence-t5-xxl & 0.684 & LVD-142M & DinoV2 & 142 & 307 & 4870 \\
    openai\_clip-vit-large-patch14-336 & All-Roberta-large-v1 & 0.677 & CLIP-400M & Lang. Supervised & 400 & 307 & 355 \\
    facebook\_dinov2-large & sentence-t5-large & 0.668 & LVD-142M & DinoV2 & 142 & 307 & 335 \\
    facebook\_dinov2-small & sentence-t5-xl & 0.661 & LVD-142M & DinoV2 & 142 & 22 & 1240 \\
    facebook\_dinov2-small & all-mpnet-base-v2 & 0.655 & LVD-142M & DinoV2 & 142 & 22 & 109 \\
    facebook\_dinov2-small & all-MiniLM-L6-v1 & 0.644 & LVD-142M & DinoV2 & 142 & 22 & 22 \\
    facebook\_convnext-base-224-22k & gtr-t5-xxl & 0.626 & ImageNet-21k & Supervised & 14.1 & 89 & 4870 \\
    timm\_vit\_small\_patch16\_224.augreg\_in1k & gtr-t5-xl & 0.602 & ImageNet-1k & Supervised & 1.2 & 22 & 1240 \\
    timm\_convnext\_base.fb\_in22k & all-MiniLM-L6-v2 & 0.590 & ImageNet-21k & Supervised & 14.1 & 89 & 22 \\
    timm\_convnext\_tiny.fb\_in1k & gtr-t5-xl & 0.540 & ImageNet-1k & Supervised & 1.2 & 29 & 1240 \\
    timm\_convnext\_base.fb\_in1k & msmarco-bert-base-dot-v5 & 0.512 & ImageNet-1k & Supervised & 1.2 & 89 & 109 \\
    facebook\_dino-vitb8 & msmarco-distilbert-dot-v5 & 0.445 & ImageNet-1k & DinoV1 & 1.2 & 86 & 66 \\
    facebook\_dino-vits8 & all-mpnet-base-v2 & 0.423 & ImageNet-1k & DinoV1 & 1.2 & 22 & 109 \\
    facebook\_dino-vits8 & paraphrase-TinyBERT-L6-v2 & 0.398 & ImageNet-1k & DinoV1 & 1.2 & 22 & 66 \\
    \bottomrule
    \end{tabular}
    }
    
\end{table*}
% \end{sidewaystable*}

\section{Qualitative results}
\label{A_sec:qual}
In Table \ref{tab:image_examples}, we present instances of retrieval mispredictions where the original image fails to rank within the top five closest images to the given caption, as determined by local Kernel CKA method. Building upon the experimental methodology outlined in the main paper, we selected 320 base samples and conducted local Kernel CKA retrieval using an additional 500 query samples. We used All-Roberta-large-v1 for text embeddings and DINOv2 ViT-L/14 for image embeddings.
The results distinctly illustrate that despite the failure to retrieve the exact original image, the alternative images identified in the top five still exhibit a considerable degree of semantic similarity to the provided caption. This underscores the robustness of the local Kernel CKA retrieval approach, revealing its capability to identify images that, while not the precise match, maintain semantic coherence with the specified caption.

\clearpage
\raggedbottom

% {
%     \small
%     \bibliographystyle{ieeenat_fullname}
%     \bibliography{main}
% }

% \input{sec/summary}

\end{document}